\documentclass[runningheads]{llncs}
\usepackage{graphicx}

\usepackage{tikz}
\usepackage{comment} 
\usepackage{amsmath,amssymb} %
\usepackage{color}

\usepackage{graphicx}
\usepackage{comment}
\usepackage{amsmath,amssymb} %
\usepackage{color}

\usepackage{xspace}
\usepackage{url}
\usepackage{caption}
\usepackage{subcaption}
\usepackage{wrapfig}
\usepackage{array}
\usepackage{float}
\usepackage{enumitem}
\usepackage{fancyhdr}

\usepackage{color,soul}
\usepackage{epsfig}
\usepackage{graphicx}
\usepackage{amsmath}
\usepackage{amssymb}
\usepackage{enumitem}
\usepackage{caption}
\usepackage{subcaption}
\usepackage{booktabs}
\usepackage[export]{adjustbox}
\usepackage{makecell}
\usepackage{mathtools}

\usepackage[pagebackref=true,breaklinks=true,letterpaper=true,colorlinks,bookmarks=false]{hyperref}

\definecolor{darkgreen}{rgb}{0,0.5,0}
\definecolor{orange}{rgb}{1,0.5,0}
\definecolor{teal}{rgb}{0,0.5,0.5}
\definecolor{darkpurple}{rgb}{0.5, 0, 0.5}
\definecolor{darkorange}{rgb}{0.8,0.4,0}

\newcommand{\etal}{et al.\xspace}

\DeclareMathOperator*{\argmin}{arg\,min}
\newcommand{\CIRCLE}[1]{\raisebox{.5pt}{\footnotesize \textcircled{\raisebox{-.6pt}{#1}}}}

\begin{document}
\pagestyle{headings}
\mainmatter
\def\ECCVSubNumber{5554}  %

\title{GeLaTO: Generative Latent Textured Objects} %

\titlerunning{GeLaTO: Generative Latent Textured Objects}
\author{Ricardo Martin-Brualla \and
Rohit Pandey \and
Sofien Bouaziz \and \\
Matthew Brown \and
Dan B Goldman}
\authorrunning{R. Martin-Brualla et al.}
\institute{Google Research\\
\email{\{rmbrualla,rohitpandey,sofien,mtbr,dgo\}@google.com}\\
Project website: \texttt{\href{https://gelato-paper.github.io}{https://gelato-paper.github.io}}\vspace{-10pt}}
\maketitle

\begin{abstract}

Accurate modeling of 3D objects exhibiting transparency, reflections and thin structures is an extremely challenging problem. Inspired by billboards and geometric proxies used in computer graphics, this paper proposes Generative Latent Textured Objects (GeLaTO), a compact representation that combines a set of coarse shape proxies defining low frequency geometry with learned neural textures, to encode both medium and fine scale geometry as well as view-dependent appearance. To generate the proxies' textures, we learn a joint latent space allowing category-level appearance and geometry interpolation. The proxies are independently rasterized with their corresponding neural texture and composited using a U-Net, which generates an output photorealistic image including an alpha map. We demonstrate the effectiveness of our approach by reconstructing complex objects from a sparse set of views. We show results on a dataset of real images of eyeglasses frames, which are particularly challenging to reconstruct using classical methods. We also demonstrate that these coarse proxies can be handcrafted when the underlying object geometry is easy to model, like eyeglasses, or generated using a neural network for more complex categories, such as cars.

\end{abstract}

\section{Introduction}

Recent research in category-level view and shape interpolation has largely focused on generative methods~\cite{Karras2019StyleGAN} due to their ability to generate realistic and high resolution images. To close the gap between generative models and 3D reconstruction approaches, we present a method that embeds a generative model in a compact 3D representation based on textured-mapped proxies. 

Texture-mapped proxies have been used as a substitute for complex geometry since the early days of computer graphics. Because manipulating and rendering geometric proxies is much less computationally intensive than corresponding detailed geometry, this representation has been especially useful to represent objects with highly complex appearance such as clouds, trees, and grass~\cite{Decoret2003BillboardClouds,Rohlf1994Billboards}. Even today, with the availability of powerful graphics processing units, real-time game engines offer geometric representations with multiple levels of detail that can be swapped in and out with distance, using texture maps to supplant geometry at lower levels of detail. 

This concept can be adapted to deep learning, for which the capacity of a network that can learn complex geometry might be larger than the capacity needed to learn its surface appearance under multiple viewpoints. Inspired by texture-mapped proxies, we propose a representation consisting of four parts: \CIRCLE{1} a 3D proxy geometry that coarsely approximates the object geometry; \CIRCLE{2} a view-dependent deep texture encoding the object's surface light field, including view-dependent effects like specular reflections, and geometry that lies away from the proxy surface; \CIRCLE{3} a generative model for these deep textures that can be used to smoothly interpolate between models, or to reconstruct unseen object instances within the category; \CIRCLE{4} a U-Net to re-render and composite all the Neural Proxies into a final RGB image and a transparency mask. 

To evaluate our approach we capture a dataset of 85 eyeglasses frames and demonstrate that our compact representation is able to generate realistic reconstructions even for these complex objects featuring transparencies, reflections and thin features. In particular, we use three planar proxies to model eyeglasses and show that using our generative model, we can reconstruct an instance with more accuracy and $3\times$ fewer input views compared to a model optimized exclusively for that instance. We also show compelling interpolations between instances of the dataset, and a prototype virtual try-on system for eyeglasses.
Finally, we qualitatively evaluate our representation on cars from the ShapeNet dataset~\cite{shapenet2015}, for which we use five free-form parameterized textured mesh proxies learnt to model car shapes~\cite{Groueix2018AtlasNet}. Supplementary video and more results are available at the project website: \texttt{\href{https://gelato-paper.github.io}{https://gelato-paper.github.io}}.

To summarize, our main contributions are: \CIRCLE{1}~a novel compact representation to capture the appearance and geometry of complex real world objects; \CIRCLE{2}~a re-rendering and compositing step that can handle transparent objects; \CIRCLE{3}~a learned latent space allowing category-level interpolation; \CIRCLE{4}~few-shot reconstruction, using a network pre-trained on a corpus of the corresponding object category.

\section{Related Work}

\begin{figure}[t]
	\centering
	\begin{subfigure}{0.32\columnwidth}
		\centering
		\includegraphics[height=18mm]{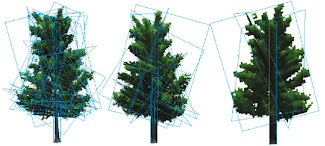}
		\caption{Traditional billboards}
	\end{subfigure}
	\begin{subfigure}{0.32\columnwidth}
		\centering
		\includegraphics[height=18mm]{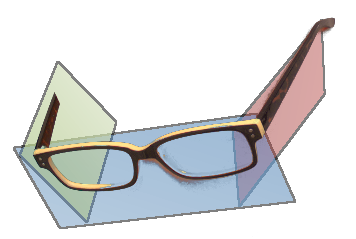}
		\caption{Planar proxies}
	\end{subfigure}
	\begin{subfigure}{0.32\columnwidth}
		\centering
		\includegraphics[height=18mm]{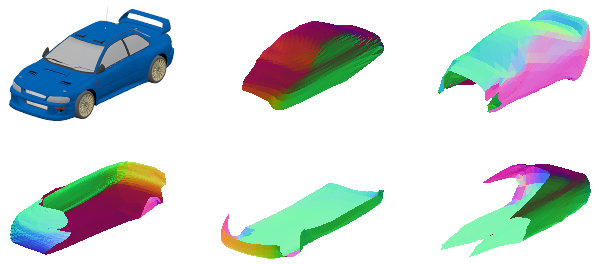}
		\caption{Free-form proxies}
	\end{subfigure}
	\caption{
	Inspired by (a) traditional computer graphics billboards~\cite{tree_billboard_figure}, our representation uses (b) planar proxies for classes with well-bounded geometric variations like eyeglasses, and (c) free-form 3D patches for generic classes like cars.
	}
   \label{fig:image-based-fusion-diagram}
\end{figure}

\subsection{3D reconstruction}

Early work in 3D reconstruction attempted to model a single object instance or static scene \cite{pollefeys2004visual} by refining multiview image correspondences~\cite{Furukawa2007PMVS} along with robust estimation of camera geometry.
These methods work well for rigid, textured scenes but are limited by assumptions of Lambertian reflectance. Later work attempts to address this, for example using active illumination to capture reflectance~\cite{Tunwattanapong2013ReflectanceAndShape}, known backgrounds to reason about transparency~\cite{Shan2012RefractiveHeightFields}, or special markers on the scanner to recognise mirrors~\cite{Wheelan2018Mirrors}. Thin structures present special challenges, which Liu~\etal~\cite{Liu2019CurveFusion} address by fusing of RGBD observations over multiple views. Even with such specifically engineered solutions, reconstruction of thin structures, reflection and transparency remain open research problems, and strong object or scene priors are desirable to enable accurate 3D reconstruction.

Recent progress in deep learning has renewed efforts to develop scene priors and object category models. Kar~\etal~\cite{Kar2015CategorySpecificObjectReconstruction} learn a linear shape basis for 3D keypoints for each category, using a variant of NRSfM~\cite{bregler2000recovering}.
Kanazawa~\etal~\cite{Kanazawa2018CMR} learn category models using a fixed deformable mesh, with a silhouette based loss function trained via a differentiable mesh renderer. Later work to regress mesh coordinates directly from the image, trained via cycle consistency, showed generalization across deformations for a class-specific mesh~\cite{Kulkarni2019CanonicalSurfaceMapping}. Chen~\etal represent view dependent effects by learning surface lightfields~\cite{Chen2018DeepSurfaceLightFields}. Implicit surface models~\cite{Chen_2019_CVPR,Mescheder2019OccNet,park2019deepsdf} use a fully connected network to represent the signed surface distance as a function of 3D coordinate.

\subsection{Neural Rendering}

Neural rendering techniques relax the requirement to produce a fully specified physical model of the object or scene, generating instead an intermediate representation that requires a neural network to render. We refer the reader to the comprehensive survey of Tewari~\etal~\cite{Tewari2020NeuralSTAR}.
Recent works use volumetric representations that can be learned on a voxel grid~\cite{lombardi2019neural,sitzmann2019deepvoxels}, or modeled directly as a function taking 3D coordinates as input~\cite{mildenhall2020nerf,sitzmann2019scene}. These methods tend to be computationally  expensive and have limited real-time performance (except for ~\cite{lombardi2019neural}).
Neural textures~\cite{Thies2019NeuralTextures} jointly learn features on a texture map along with a U-Net. IGNOR~\cite{Thies2018IGNOR} incorporates view dependent effects by modelling the difference between true appearance and a diffuse reprojection. Such effects are difficult to predict given the scene knowledge, so GAN based loss functions are often used to render realistic output. Deep Appearance Models~\cite{lombardi2018deep_appearance_models} use a conditional variational autoencoder to generate view-dependent texture maps of faces. Image-to-image translation (pix2pix)~\cite{Isola2017Pix2pix} is often used as a general baseline. 
HoloGAN learns a 3D object representation such that sampled reprojections under a transform fool a discriminator~\cite{NguyenPhuoc2019HoloGAN}.  
Point-cloud representations are also popular for neural rerendering~\cite{Meshry2019NeuralRerenderingWild,pittaluga2019revealing} or to optimize neural features on the point cloud itself~\cite{aliev2019neuralpointbasedgraphics}.

\section{Generative Latent Textured Objects}

Our representation is inspired by proxy geometry used in computer graphics. We encode the geometric structure using a set of coarse proxy surfaces shown in Figure~\ref{fig:image-based-fusion-diagram}, and shape, albedo, and view dependent effects using view-dependent neural textures. The neural textures are parameterized using a generative model that can produce a variety of shape and appearances.

\begin{figure}[t]
	\centering
		\includegraphics[width=\columnwidth]{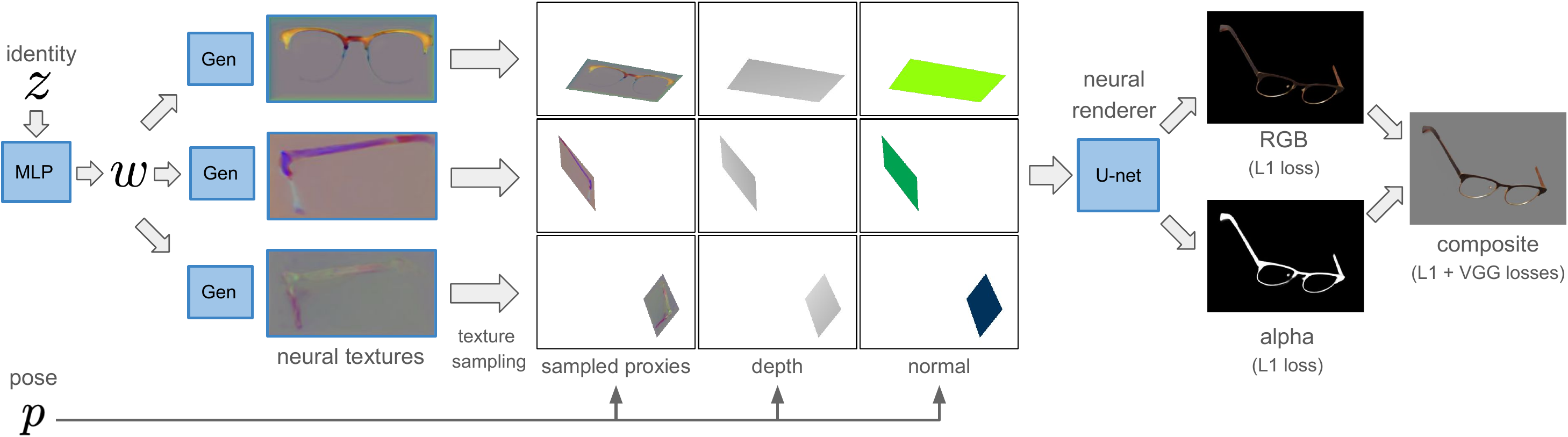}
	\caption{Network architecture. See Section~\ref{sec:architecture} for details.}
   \label{fig:network-diagram}
\end{figure}

\subsection{Model}
\label{sec:neural-proxies}
 
Given a collection of objects of a particular class, we define a latent code for each instance $i$ as $\mathbf{z}_i \in \mathbb{R}^n$. We assume that a coarse geometry consisting of a set of $K$ proxies $\{P_{i,1}, \hdots, P_{i,K}\}$, i.e. triangular meshes with UV-coordinates, is available. Our network computes a neural texture $T_{i,j} = \text{Gen}_{j}(\mathbf{w}_i)$ for each instance and proxy, where $\mathbf{w}_i = \text{MLP}(\mathbf{z}_i)$ is a non-linear reparametrization of the latent code $\mathbf{z}_i$ using an MLP. The image generators $\text{Gen}_{j}(\cdot)$ are decoders, that take a latent code as input and generate a feature map. To render an output view, we rasterize a deferred shading deep buffer from each proxy consisting of the depth, normal and UV coordinates. We then sample the corresponding neural texture using the deep buffer UV coordinates for each proxy. The deep buffers are finally processed by a U-Net~\cite{Ronneberger2015Unet} that generates four output channels, three color channels interpreted as color premultiplied by alpha~\cite{Porter1984Compositing}, and a separate alpha channel. We use color values premultiplied by alphas because color in pixels with low alpha tends to be particularly noisy in the extracted mattes and distracts the network when using reconstruction losses on the RGB components.

\subsection{Training and Architecture Details}
\label{sec:architecture}

Our network architecture is depicted in Figure~\ref{fig:network-diagram}. We use the Generative Latent Optimization (GLO) framework~\cite{GLO} to train our network end to end using simple $\ell_1$ and perceptual reconstruction losses~\cite{Johnson2016Perceptual}. We use reconstruction $\ell_1$ losses on the premultiplied RGB values, alphas, and a composite on a neutral gray background. We also apply a perceptual loss on the composite using the 2nd and 5th layers of VGG pretrained on ImageNet~\cite{deng2009imagenet}. We found adversarial losses lead to worse results, and we apply no regularization losses on the latent codes. 

The latent codes $\mathbf{z}$ for each class are randomly initialized, and we use the Adam~\cite{kingma2014adam} optimizer with a learning rate of $1e^{-5}$. We use neural textures of $9$ channels, and $\mathbf{z}$ and $\mathbf{w}$ are $8$ and $512$ dimensions respectively.  We generate results at a $512\times512$ resolution for the eyeglasses dataset and $256\times 256$ for ShapeNet. The latent transformation MLP has 4 layers of $256$ features, and the rendering U-Net contains 5 down- and up-sampling blocks with 2 convolutions each, and uses BlurPool layers~\cite{zhang2019making}, see more details in the supplementary.

\section{Dataset}

The \emph{de facto} standard for evaluating category-level object reconstruction approaches is the ShapeNet dataset~\cite{shapenet2015}. Shapenet objects can be rendered under different viewpoints, generating  RGB images with ground truth poses, and masks for multiple objects of the same category. 

Although using a synthetic dataset can help in analyzing 3D reconstruction algorithms, synthetically rendered images do not capture the complexities of real-world data. To evaluate our approach we acquire a challenging dataset of eyeglasses frames. We choose this object category because eyeglasses are physically small and have well-bounded geometric variations, making them easy to photograph under controlled settings, but they still exhibit complex structures and materials, including transparency, reflections, and thin geometric features.

\subsection{Eyeglasses Frames}
\begin{figure}[t]
   \footnotesize
	\centering
	\begin{subfigure}{0.225\columnwidth}
		\centering
		\includegraphics[height=24.5mm]{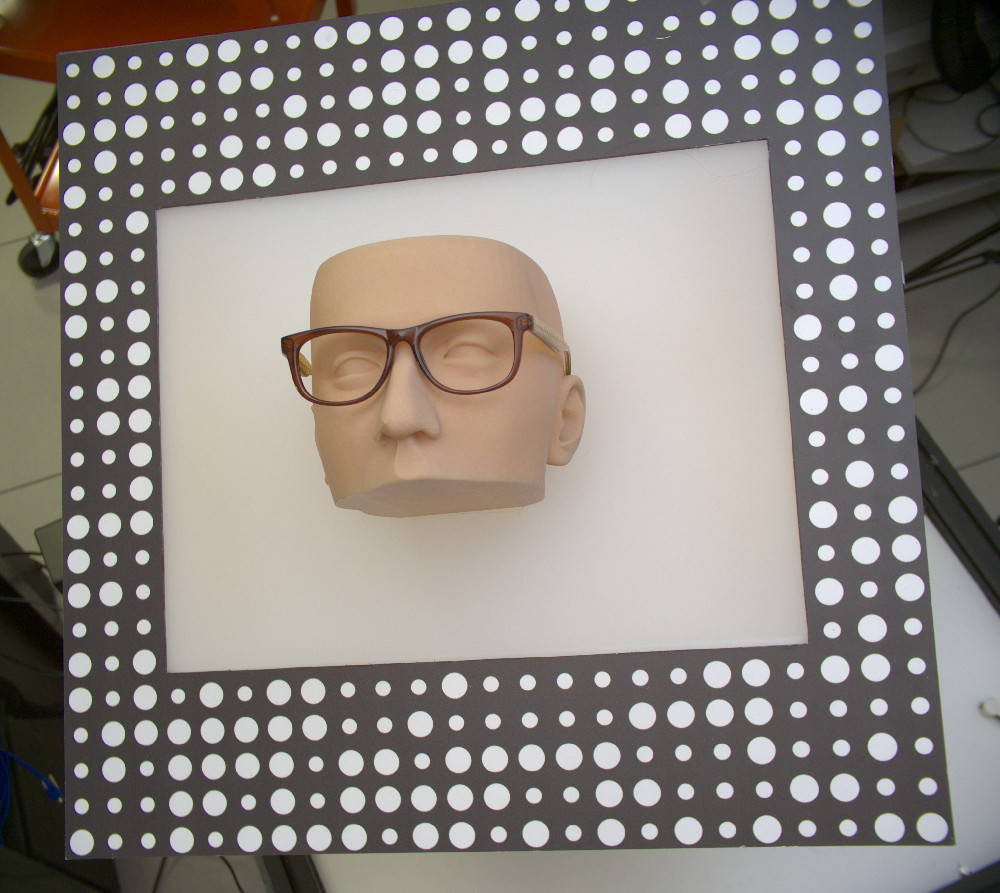}
		\caption{fixture}\label{fig:dataset_a}
	\end{subfigure}
	\begin{subfigure}{0.22\columnwidth}
		\centering
		\includegraphics[height=12mm]{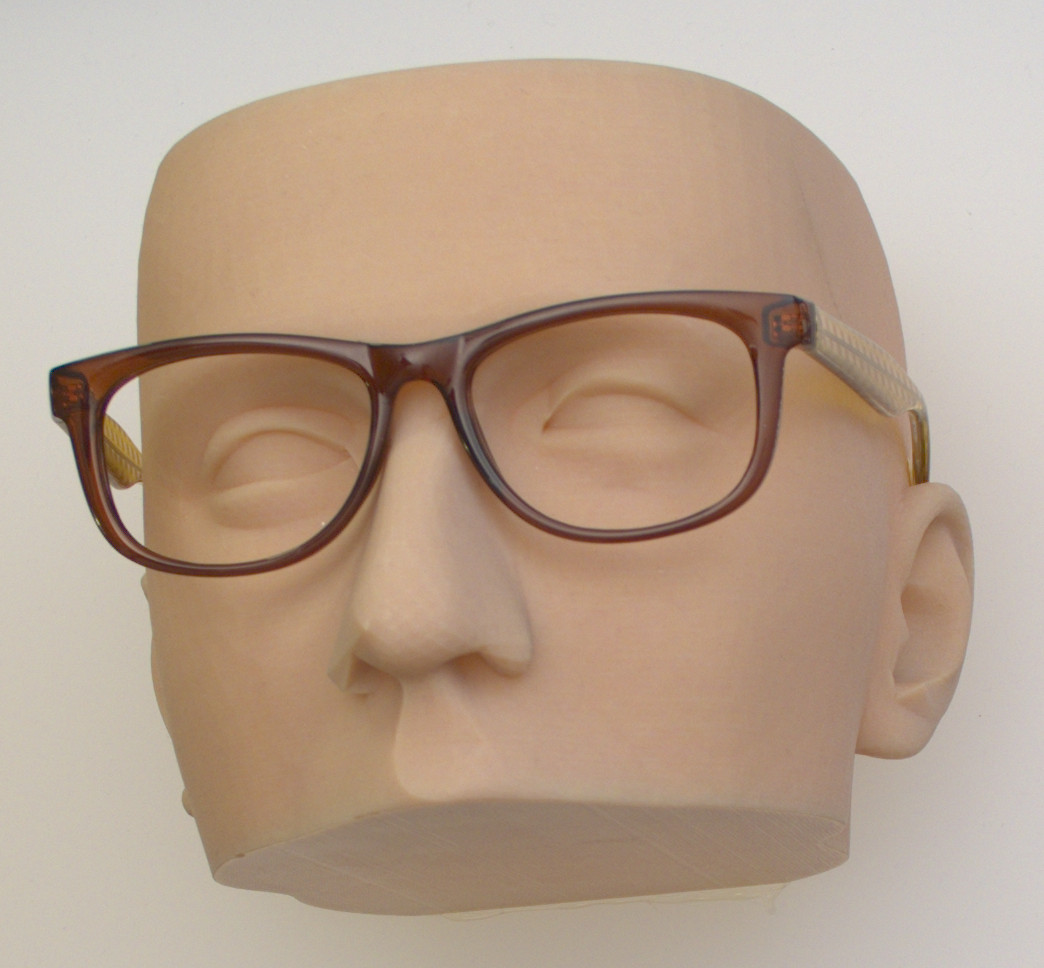}\hspace{1.5pt}\includegraphics[height=12mm]{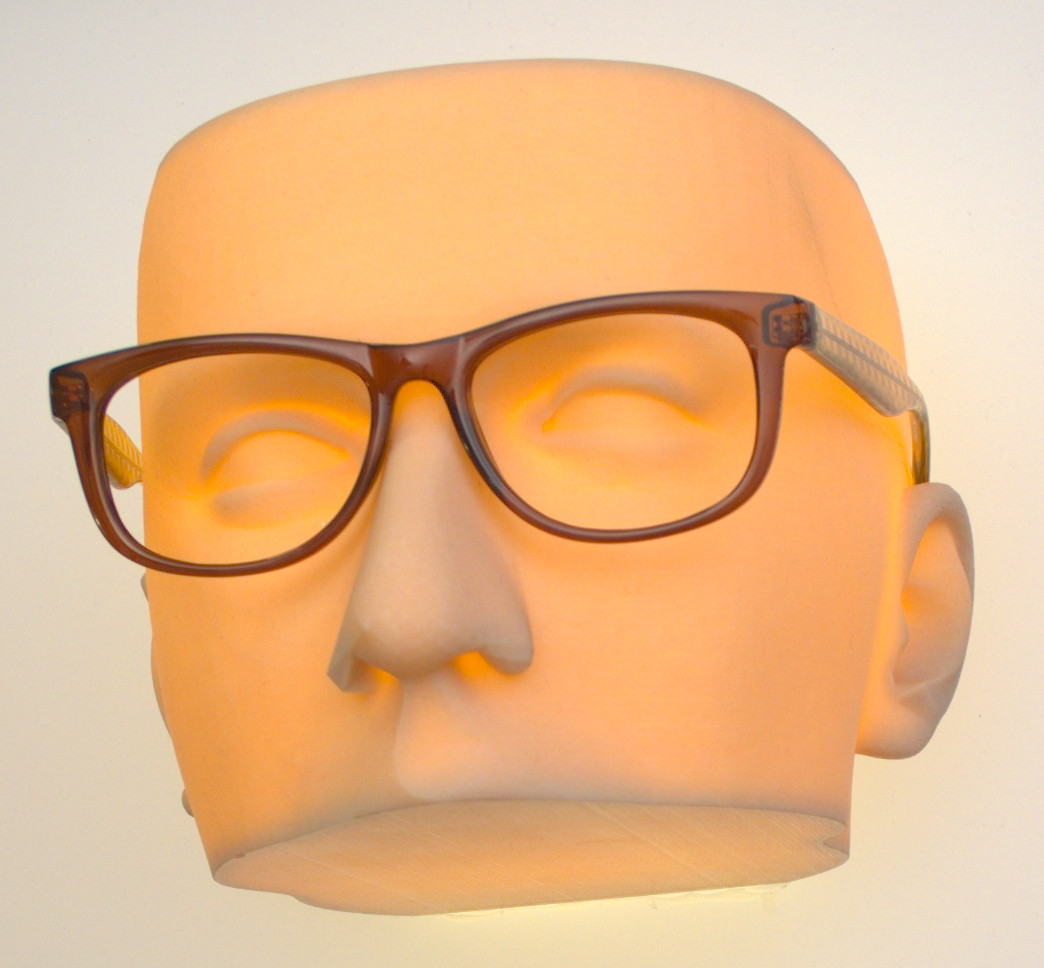}\\
		\includegraphics[height=12mm]{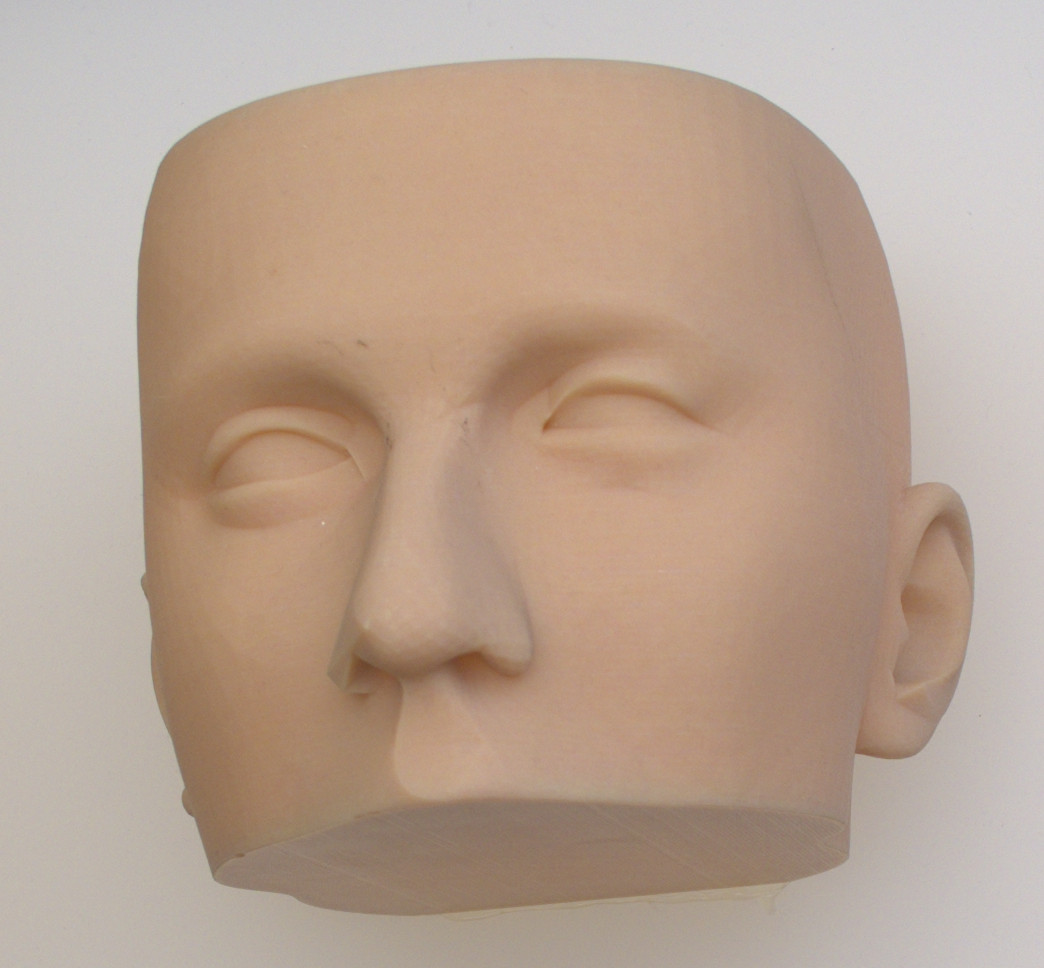}\hspace{1.5pt}\includegraphics[height=12mm]{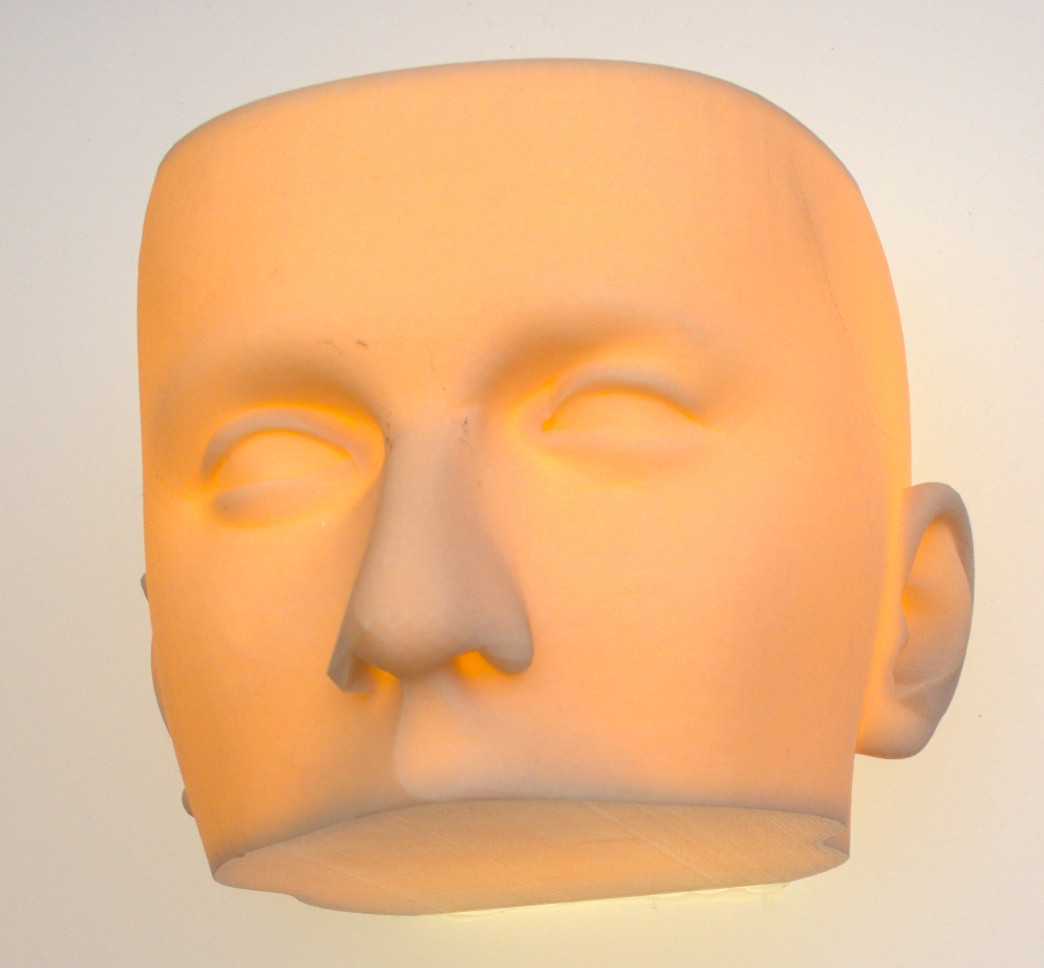}
		\caption{captures}\label{fig:dataset_b}
	\end{subfigure}
	\begin{subfigure}{0.47\columnwidth}
		\centering
		\includegraphics[height=12mm]{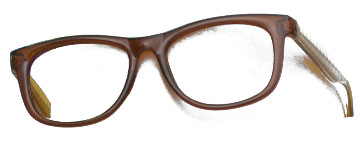}\hspace{1.5pt}%
		\includegraphics[height=12mm]{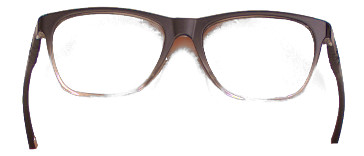}\\
		\includegraphics[height=12mm]{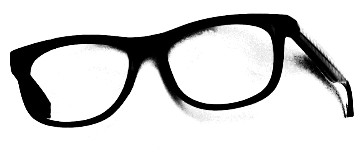}\hspace{1.5pt}%
		\includegraphics[height=12mm]{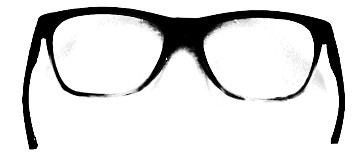}\\
		\caption{extracted RGB and alpha}\label{fig:dataset_c}
	\end{subfigure}
	\caption{
	(a) Our capture fixture includes a backlit mannequin head and white acrylic plate, surrounded by a Calibu calibration pattern~\cite{Calibu}, all of which are actuated by a robot arm.  We capture (b) four conditions for each pose and object, and solve for (c) foreground alpha mattes and colors. Note some shadows of the eyeglasses remain unmasked, due to limitations of the matting approach.}
\end{figure}

We collect a dataset of 85 eyeglasses frames under different viewpoints and fixed illumination. To capture the frames, we design a robotic fixture to sample $24\times24$ viewpoints spanning approximately $\pm24$ degrees in yaw and azimuth (Figure~\ref{fig:dataset_a}). The fixture includes a Calibu pattern~\cite{Calibu} with 3 vertical and 5 horizontal rows, enabling accurate pose estimation. The fixture center features a hollow 3D printed mannequin head and contains a light inside. For each pose, we capture an image with this backlight on and off (Figure~\ref{fig:dataset_b}). We perform difference matting by subtracting the backlit images -- which contain fewer shadows -- from a reference backlit frame without glasses. We then solve for foreground and background using the closed-form matting approach of Levin~\etal~\cite{LevinLW08} (Figure~\ref{fig:dataset_c}). The robot's pose is repeatable within $0.5$ pixels, enabling precise difference matting.

We generate 3 planar billboards to model each eyeglasses instance: front, left and right. We first compute a coarse visual hull for each object using the extracted alpha masks. We then specify a region of interest in axis-aligned head coordinates, and extract a plane that best matches the surface seen from the corresponding direction. See the supplementary for a more detailed description. We use 5 instances for testing few-shot reconstruction and train on the rest.

Note that this dataset contains two types of artifacts due to the simple acquisition setup: \CIRCLE{1} shadows cast by the glasses onto the 3D head pollute the alpha mattes and RGB images; \CIRCLE{2} depending of the viewpoint, the 3D head can occlude part of the glasses frames, resulting in missing temples. We find however that these artifacts do not affect the overall evaluation of our approach.

\subsection{ShapeNet}

We also train GeLaTO using cars from ShapeNet~\cite{shapenet2015}. We generate the proxies using the auto-encoder version of AtlasNet~\cite{Groueix2018AtlasNet} which takes as input a point cloud. We train a $5$ patches/proxies model generating triangular meshes based on a $24 \times 24$ uniform grid sampling. Note that the proxies generated by AtlasNet can overlap, but our model is robust thanks to the U-Net compositing step.

\begin{table*}[t]
\centering

\setlength{\tabcolsep}{4pt}
\scriptsize		

\begin{tabular}{l|ccc|ccc}
\toprule
\multicolumn{1}{c}{}& \multicolumn{3}{c}{view interpolation} & \multicolumn{3}{c}{few-shot reconstruction} \\
\cmidrule(lr){2-4} \cmidrule(lr){5-7}
Model     &   VAE  &  DNR & Ours &   VAE  &  DNR  & Ours  
\vspace{1pt}\\
\hline
\vspace{1pt} $\text{PSNR}$ & 39.70  & 41.21 & 41.32  & 35.59 & 36.14 & 37.19 \\
$\text{PSNR}_M$ & 21.79 & 23.29 & 23.42 & 17.94 & 18.65 & 19.64\\
SSIM & 0.9897 & 0.9916 & 0.9917 & 0.9793&  0.9819 &  0.9842 \\
Mask IoU & 0.9379 &  0.9556 & 0.9556 &  0.8686 & 0.8725 & 0.9012\\
\bottomrule
\end{tabular}

\caption{
Ablation study comparing multiple baselines on view interpolation of seen instances, and of few-shot reconstruction using $N=3$ input views, where we fine-tune the whole network together with the latent code. The VAE model is inferior in both tasks, and our approach improves upon DNR in few-shot reconstruction because our textured proxies are not masked by z-buffering.
}
\label{tab:ablation_study}
\end{table*}

\section{Evaluation}

We evaluate GeLaTO on a number of tasks on the eyeglasses dataset, and then show qualitative results on ShapeNet cars. We compare our representation against baselines inspired by neural textures~\cite{Thies2019NeuralTextures} using the same proxy geometry. In particular, we modify deferred neural rendering (DNR) in two ways: we parameterize the texture using a generator network, without loss of performance, and concatenate deep buffer channels consisting of normal and depth information to the sampled neural texture, instead of multiplying the sampled neural texture by the viewing direction vector. A key difference of our method is that Thies~\etal render a deferred rendering buffer with \emph{z-buffering} before the U-Net, whereas our method \emph{stacks} the deferred rendering buffers of each texture proxy before the U-Net. Thus our network is able to ``see through'' transparent layers to other surfaces behind the frontmost proxy. We evaluate a second baseline that uses a Variational Auto-Encoder (VAE)~\cite{kingma2013auto} instead of GLO~\cite{GLO} to model the distribution of instances, where the encoder is a MLP that takes as input a one-hot encoding of the instance id (more details in the supplementary).

\begin{figure}[t]
	\centering
    \setlength{\tabcolsep}{0.0em} %
    \hspace{-10pt}
    \begin{tabular}{r m{2.22cm} m{2.22cm} m{2.22cm} m{2.22cm} m{2.22cm}}
    VAE~ &\includegraphics[height=1.20cm]{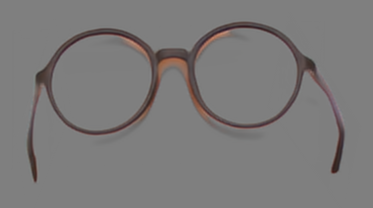} & 
    	\includegraphics[height=1.20cm]{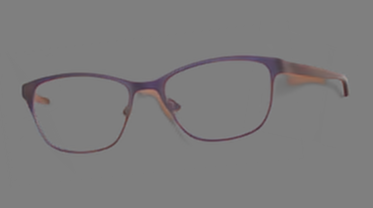} &
    	\includegraphics[height=1.20cm]{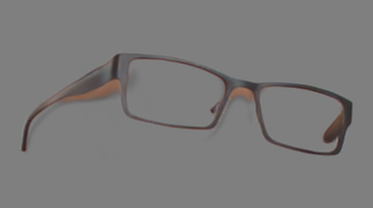} &
    	\includegraphics[height=1.20cm]{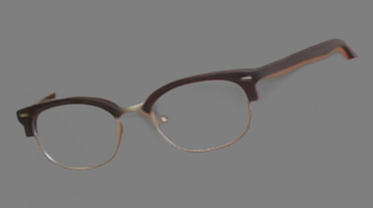} &
    	\includegraphics[height=1.20cm]{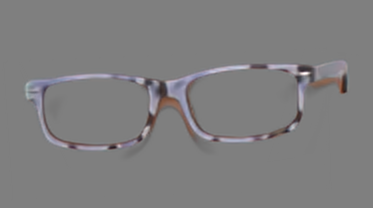} \\ 
    DNR~ & \includegraphics[height=1.20cm]{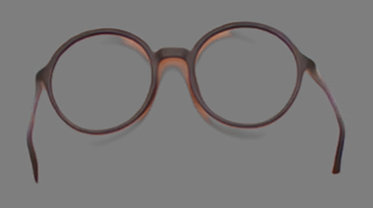} & 
    	\includegraphics[height=1.20cm]{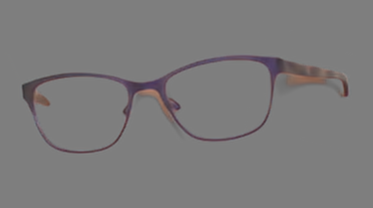} &
    	\includegraphics[height=1.20cm]{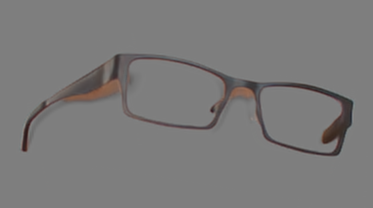} &
    	\includegraphics[height=1.20cm]{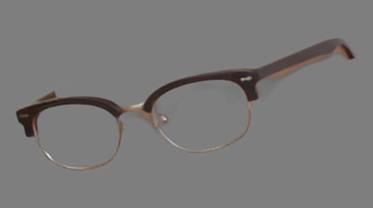} &
    	\includegraphics[height=1.20cm]{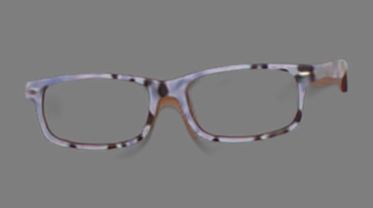} \\ 
    Ours~ & \includegraphics[height=1.20cm]{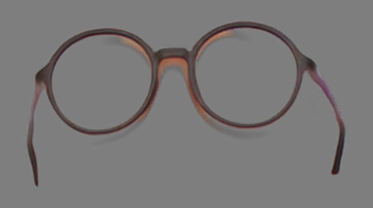} & 
    	\includegraphics[height=1.20cm]{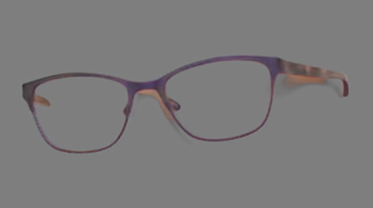} &
    	\includegraphics[height=1.20cm]{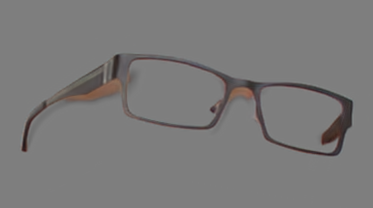} &
    	\includegraphics[height=1.20cm]{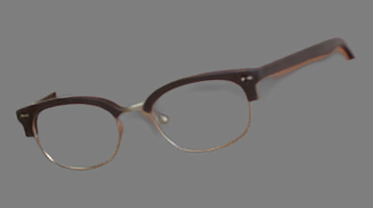} &
    	\includegraphics[height=1.20cm]{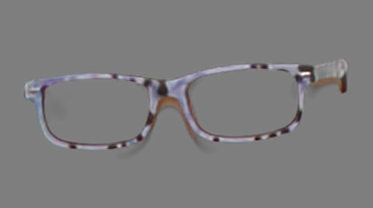} \\ 
    GT~ & \includegraphics[height=1.20cm]{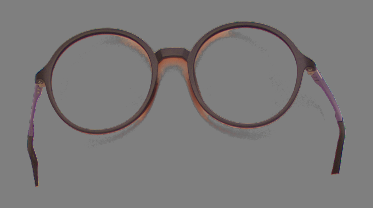} & 
    	\includegraphics[height=1.20cm]{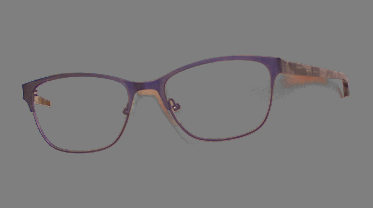} &
    	\includegraphics[height=1.20cm]{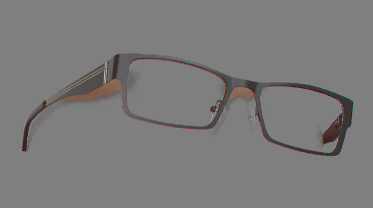} &
    	\includegraphics[height=1.20cm]{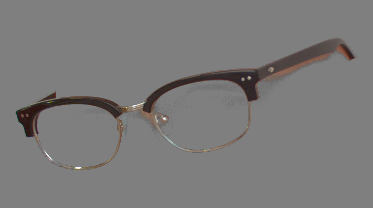} &
    	\includegraphics[height=1.20cm]{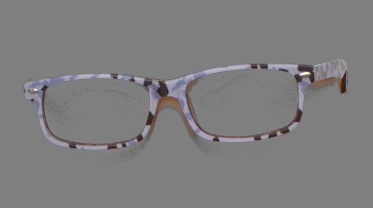}
    \end{tabular}
	\caption{
	Comparison of view interpolation results for our model and the baselines. 
	}
   \label{fig:view_interpolation_comparison}
\end{figure}
\begin{figure}[t]
	\centering
	\includegraphics[width=0.95\columnwidth]{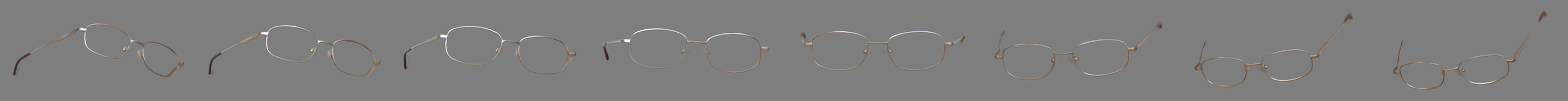}\\
	\vspace{-1pt}\includegraphics[width=0.95\columnwidth]{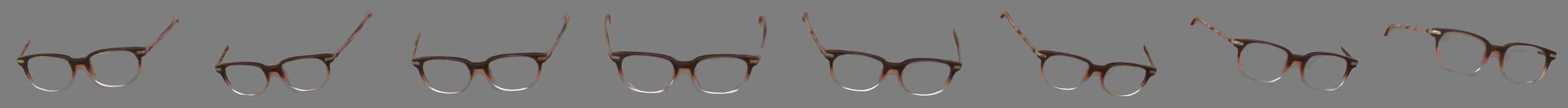}\\ 
	\vspace{-1pt}\includegraphics[width=0.95\columnwidth]{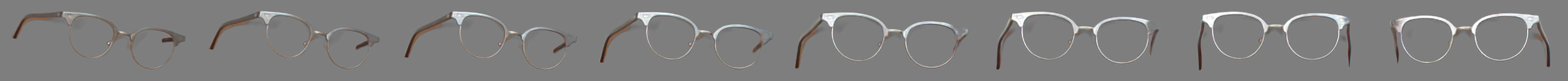}\\ 
	\vspace{-1pt}\includegraphics[width=0.95\columnwidth]{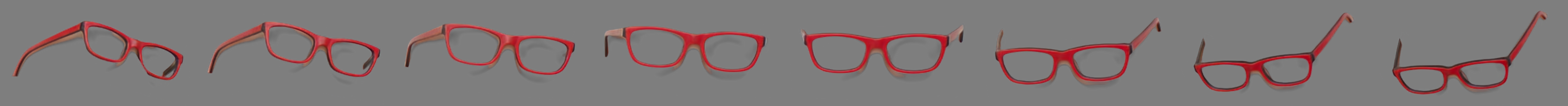}\\ 
	\vspace{-1pt}\includegraphics[width=0.95\columnwidth]{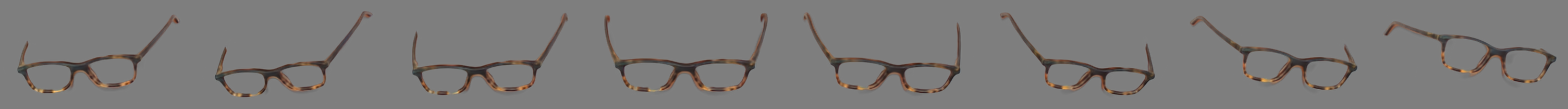}\\ 
	\vspace{-1pt}\includegraphics[width=0.95\columnwidth]{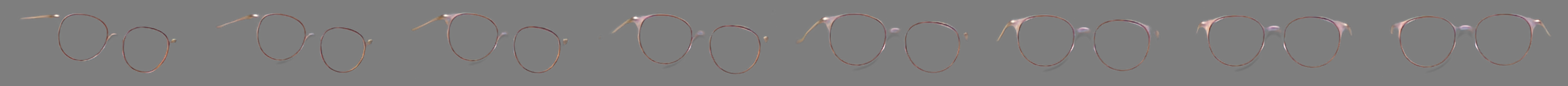}
	\caption{
	View interpolation results from our model for a variety of glasses. 
	}
   \label{fig:view_interpolation}
\end{figure}

\subsection{View Interpolation}
\label{sec:view-interpolation}

We first evaluate our method on the view interpolation task, and show that textured proxies can model complex geometry and view-dependent effects. We train a network on $98\%$ of the views of the training set of the eyeglasses dataset, and test on the remaining $2\%$. Quantitative results in Table~\ref{tab:ablation_study} show that our model slightly improves upon the DNR baseline, and is significantly better than VAE. We report PSNR and SSIM on the whole image, $\text{PSNR}_M$ evaluated within $7$ pixels of alpha $>0.1$ values, and IoU of the alpha channel thresholded at $0.5$.

Figure~\ref{fig:view_interpolation_comparison} qualitatively compares the view interpolation results. VAE results are overly smoothed, and our approach captures more high-frequency details compared to DNR.
Figure~\ref{fig:view_interpolation} contains interpolations of the eyeglasses seen from multiple viewpoints, showcasing strong view-dependent effects due to shiny or metallic metallic materials, and reconstructions of transparent glasses that are predominantly composed of specular reflections (last example).

\subsection{Instance Interpolation}

\begin{figure}[t]
	\centering
    \setlength{\tabcolsep}{0.0em} %
    \hspace{-10pt}
    \begin{tabular}{r m{0.92\columnwidth}}
    VAE~ &\includegraphics[width=0.92\columnwidth]{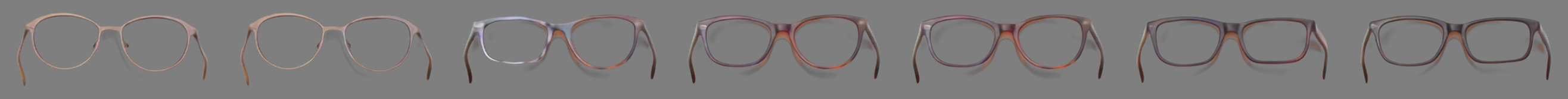}\\
    \vspace{-14pt}\\
    Ours~ &\includegraphics[width=0.92\columnwidth]{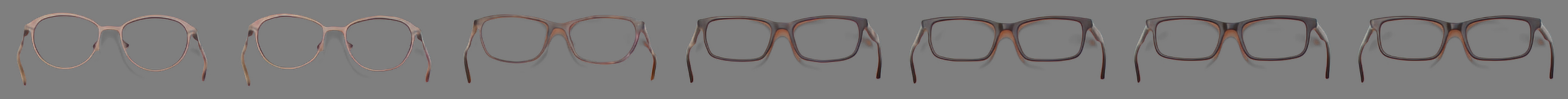}\\
    VAE~ &\includegraphics[width=0.92\columnwidth]{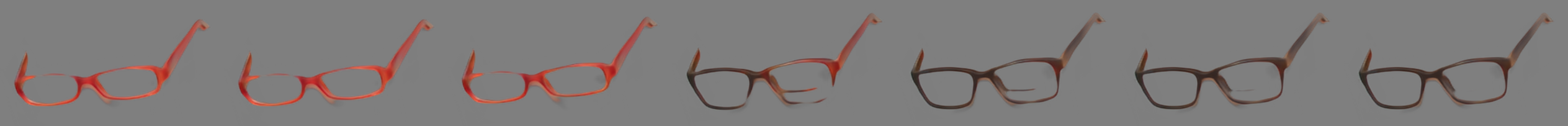}\\
    \vspace{-14pt}\\
    Ours~ &\includegraphics[width=0.92\columnwidth]{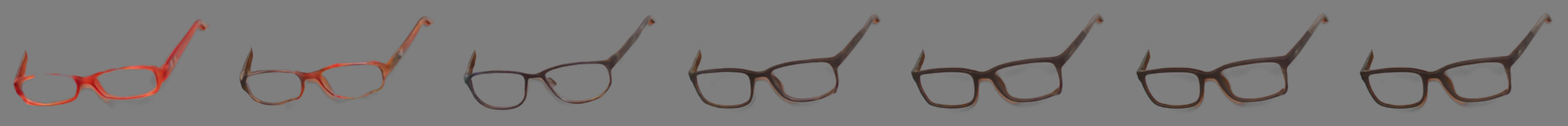}\\
    VAE~ &\includegraphics[width=0.92\columnwidth]{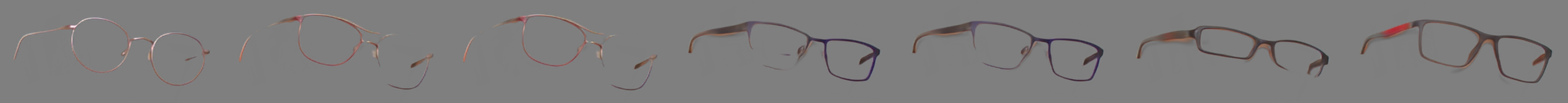}\\
    \vspace{-14pt}\\
    Ours~ &\includegraphics[width=0.92\columnwidth]{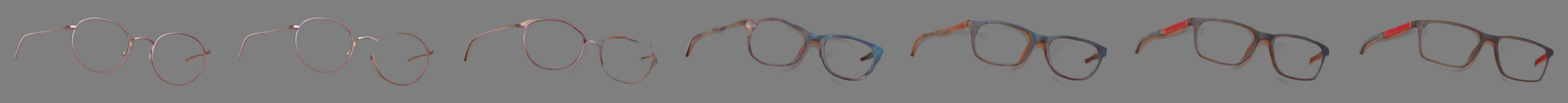}\\
    VAE~ &\includegraphics[width=0.92\columnwidth]{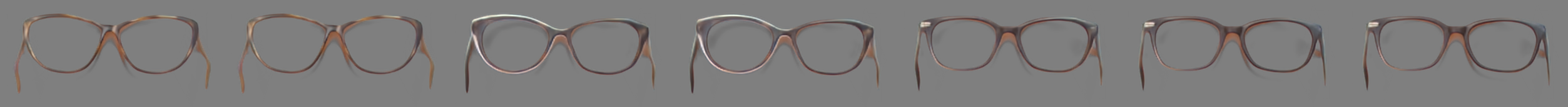}\\
    \vspace{-14pt}\\
    Ours~ &\includegraphics[width=0.92\columnwidth]{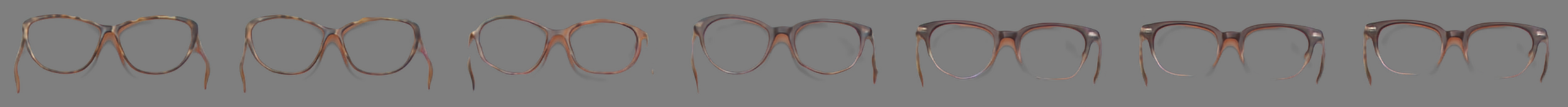}\\
    \end{tabular}
	\caption{
	Examples of instance interpolation of VAE and our model using GLO.
	}
   \label{fig:instance_interpolation}
\end{figure}

Our generative model allows interpolations in the latent space of objects, effectively building a deformable model of shape and appearance, reminiscent of 3D morphable models~\cite{blanz1999morphable}. We visualize such interpolations in Figure~\ref{fig:instance_interpolation}, in which the latent code $\mathbf{z}$ is linearly interpolated while the proxy geometry is kept constant. 
VAE models are commonly thought to have better interpolation abilities than GLO, because the injected noise regularizes the latent space. However, we find GLO offers better interpolations in our setup. VAE interpolations tend to be less visually monotonic, like in the last example where a white border appears and then disappears on the left side of the frame, and often contain spurious structures like the double rim on the second example. The supplementary video shows the effects of interpolating the neural texture and proxy geometry independently.

\begin{figure}[t]
	\centering
	\begin{tabular}{cccc}
    \includegraphics[height=1.25cm]{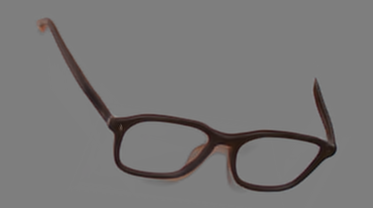} &
    \includegraphics[height=1.25cm]{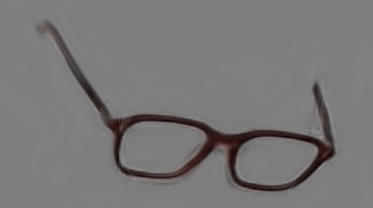} & 
    \includegraphics[height=1.25cm]{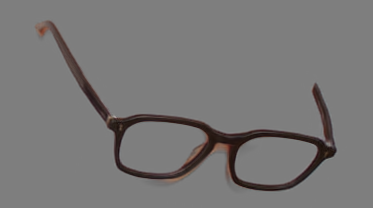} & 
    \includegraphics[height=1.25cm]{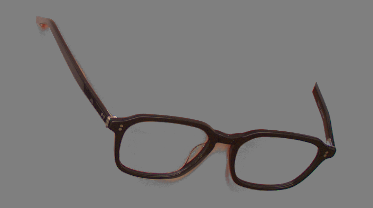} \\  
    \includegraphics[height=1.25cm]{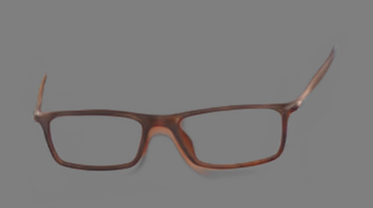} &
    \includegraphics[height=1.25cm]{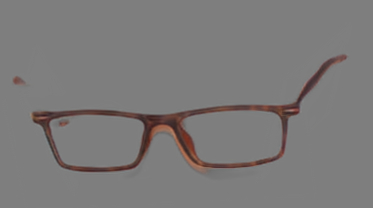} & 
    \includegraphics[height=1.25cm]{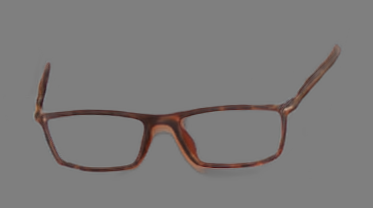} & 
    \includegraphics[height=1.25cm]{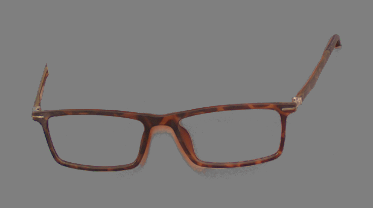} \\ 
    VAE & DNR & Ours & GT
    \end{tabular}
	\caption{
	Comparison of few-shot reconstruction using $N=3$ input views.
	}
   \label{fig:few_shot_reconstruction_comparison}
\end{figure}

\begin{figure}[t]
	\centering
	\begin{tabular}{cc}
    \includegraphics[height=3.4cm]{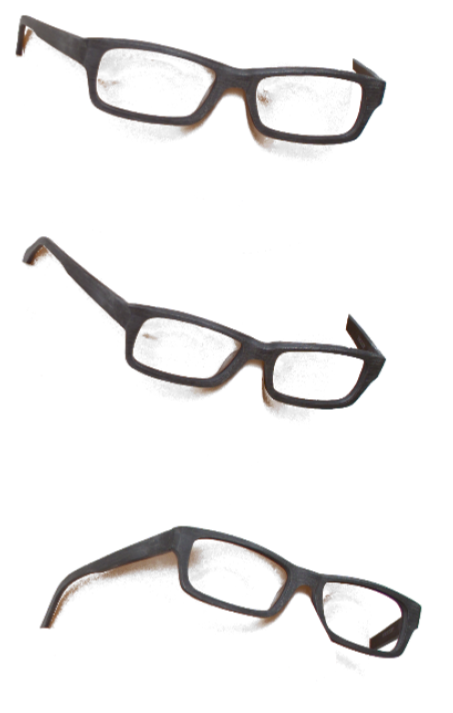} & 
    \includegraphics[height=3.6cm]{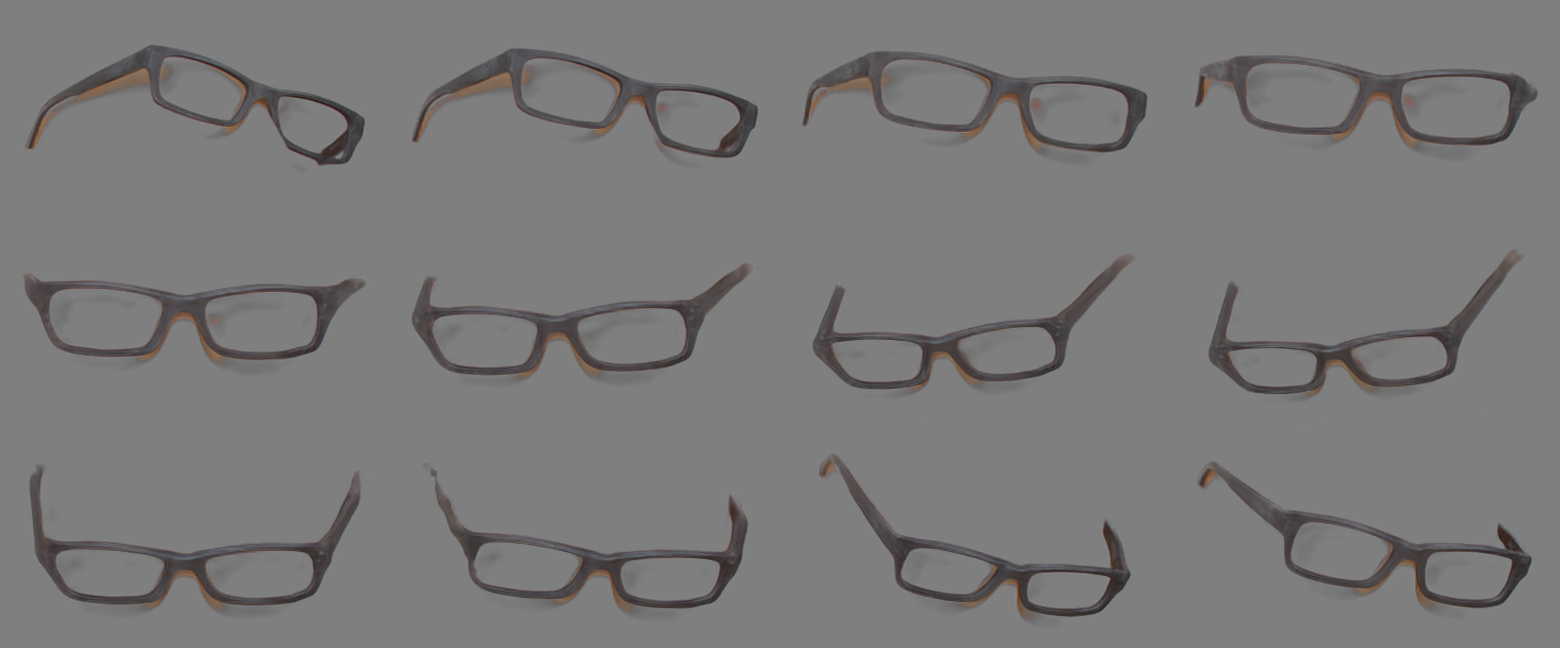} \\
    \includegraphics[height=3.4cm]{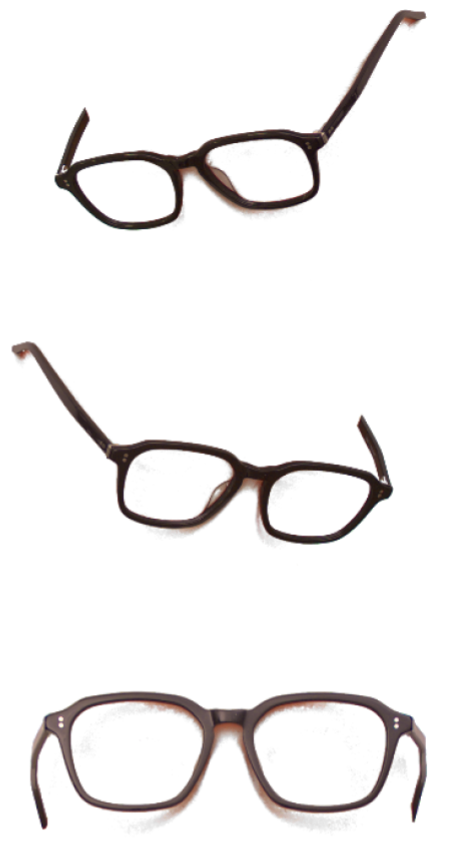} & 
    \includegraphics[height=3.6cm]{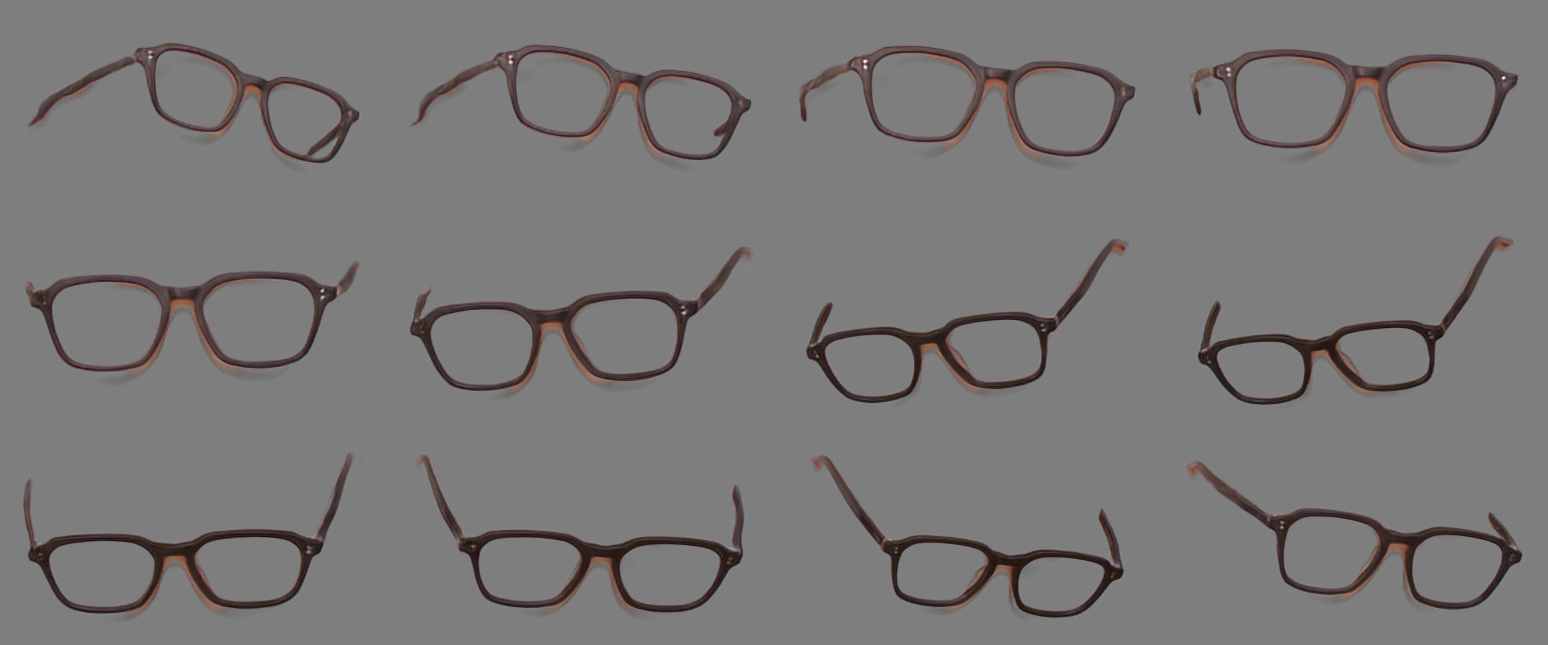} \\  
    inputs & reconstructed views \\
    
    \end{tabular}
	\caption{
	Results for few-shot reconstruction using $N=3$ views. Left: Input views. Right: Reconstructed views using our method after fine-tuning on the input views. Notice that although the first instance is only captured from the left, our network still is able to reconstruct other viewpoints effectively. We are also able to capture view-dependent effects as seen on the bridge region of the glasses.
	}
   \label{fig:few_shot_reconstruction}
\end{figure}

\begin{figure}[h]
	\centering
	\small
    \setlength{\tabcolsep}{0.5pt}
    \hspace{-5pt}
	\begin{tabular}{ccccccc}
    $N=30$ & $N=100$ & $N=3$ & $N=10$ & $N=30$ & $N=100$ & GT \\

	\includegraphics[width=0.138\columnwidth]{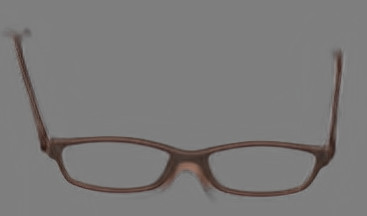} &
	\includegraphics[width=0.138\columnwidth]{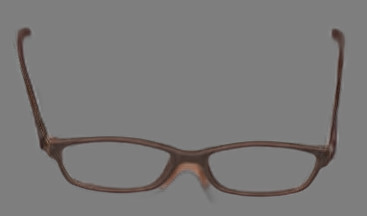} & 
	\includegraphics[width=0.138\columnwidth]{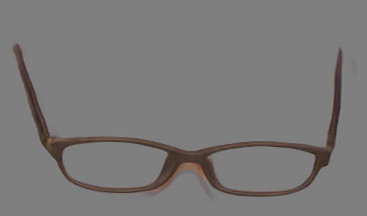} &
	\includegraphics[width=0.138\columnwidth]{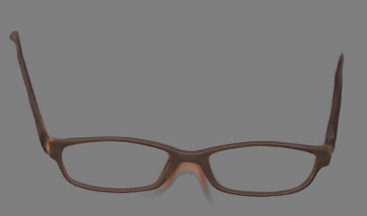} &
	\includegraphics[width=0.138\columnwidth]{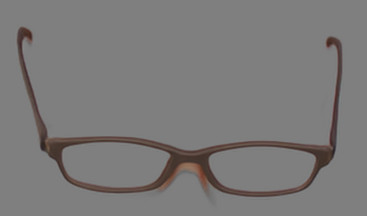} & 
	\includegraphics[width=0.138\columnwidth]{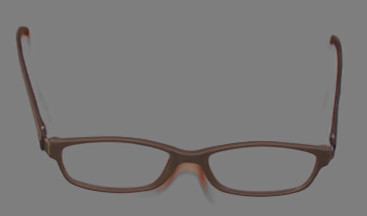} &
	\includegraphics[width=0.138\columnwidth]{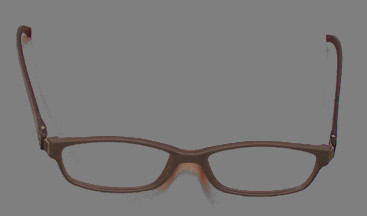}\\
	\vspace{-14pt}\\
	\includegraphics[width=0.138\columnwidth]{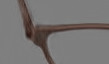} &
	\includegraphics[width=0.138\columnwidth]{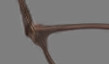} &
	\includegraphics[width=0.138\columnwidth]{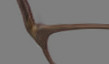} &
	\includegraphics[width=0.138\columnwidth]{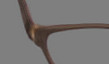} &
	\includegraphics[width=0.138\columnwidth]{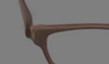} & 
	\includegraphics[width=0.138\columnwidth]{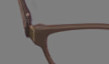} &
	\includegraphics[width=0.138\columnwidth]{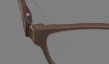}\\
    \multicolumn{2}{c}{DNR from scratch} & \multicolumn{4}{c}{finetuning category-level model} \\

     &  & $N=3$ & $N=10$ & $N=30$ & $N=100$ & \\
    & & 
    \includegraphics[width=0.138\columnwidth]{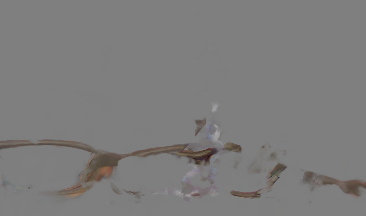} &
	\includegraphics[width=0.138\columnwidth]{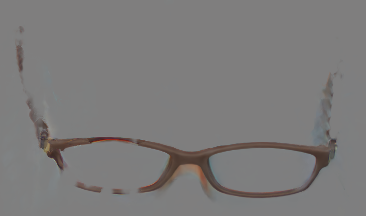} &
	\includegraphics[width=0.138\columnwidth]{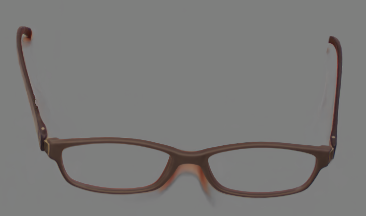} & 
	\includegraphics[width=0.138\columnwidth]{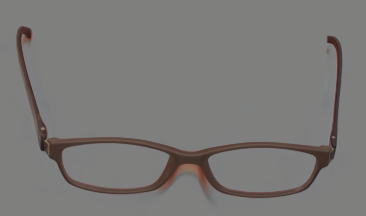}\\
    & & 
    \includegraphics[width=0.138\columnwidth]{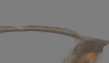} &
	\includegraphics[width=0.138\columnwidth]{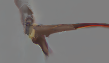} &
	\includegraphics[width=0.138\columnwidth]{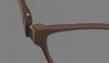} & 
	\includegraphics[width=0.138\columnwidth]{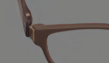}\\
    & & \multicolumn{4}{c}{neural radiance fields} 

    \end{tabular}

	\caption{
	Unseen instance reconstruction varying the number of input images $N$.
	}
   \label{fig:num_images}
\end{figure}

\subsection{Few-shot reconstruction}

Because we have parameterized the space of textures, we can think of reconstructing a particular instance by finding the right latent code $\mathbf{z}$ that reproduces the input views. This can be done either using an encoder network, or by optimization via gradient descent on a reconstruction loss. These approaches are unlikely to yield good results in isolation, because the dimensionality of the object space can be arbitrarily large compared to the dimensionality of the latent space, e.g., when objects exhibit a print of a logo or text. As noted by Abdal~\etal~\cite{Abdal2019Image2StyleGAN}, optimizing intermediate parameters of the networks instead can yield better results, like the transformed latent space $\mathbf{w}$, the neural texture space, or even optimizing all the network parameters, i.e. fine-tuning the whole network.

Thus, given a set of views $\{I_1, \hdots, I_k\}$ with corresponding poses $\{\mathbf{p}_1 \hdots \mathbf{p}_k\}$ and proxy geometry $\{P_{1}, \hdots, P_{K}\}$, we define a new latent code $\mathbf{z}$ and set the reconstruction process as optimization
\[
\mathbf{z}^{\star}, \boldsymbol{\theta}^{\star} = \argmin_{\mathbf{z}, \boldsymbol{\theta}}\sum_k \|I^k - \text{Net}(\mathbf{z}, \mathbf{p}_k, \boldsymbol{\theta})\|_1,
\]
where $\text{Net}(\cdot, \cdot, \cdot)$ is the end to end network depicted in Figure~\ref{fig:network-diagram} parameterized by the latent code $\mathbf{z}$, the pose $\mathbf{p}$, and the network parameters $\boldsymbol{\theta}$.

\begin{figure}[]

    \setlength{\tabcolsep}{0.4pt}

	\centering
	\begin{subfigure}[t]{0.20\columnwidth}
    	\centering
    	\quad\\
        \vspace{-6em}
    	\includegraphics[width=\columnwidth]{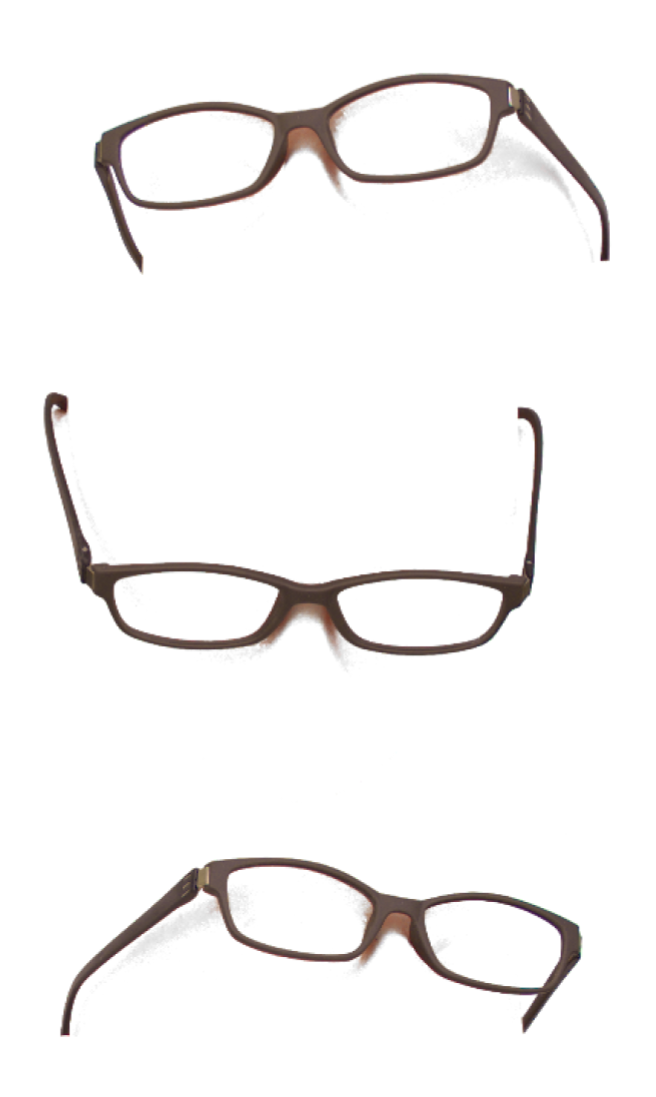}\\
    	inputs
    \end{subfigure}
	\begin{subfigure}[t]{0.22\columnwidth}
    	\centering
    	\includegraphics[width=\columnwidth]{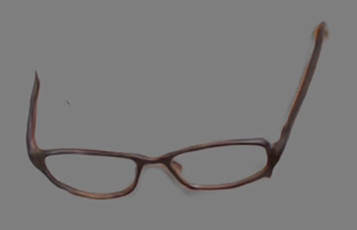}\\
    	$z$\\
    	\vspace{2pt}
    	\includegraphics[width=\columnwidth]{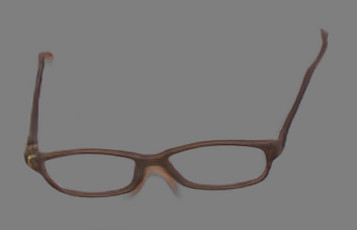}\\
    	texgen
    \end{subfigure}
	\begin{subfigure}[t]{0.22\columnwidth}
    	\centering
    	\includegraphics[width=\columnwidth]{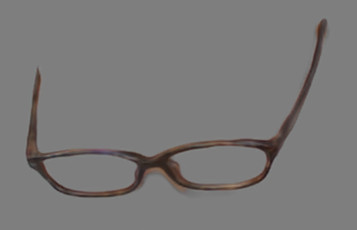}\\
    	$w$\\
    	\vspace{2pt}
    	\includegraphics[width=\columnwidth]{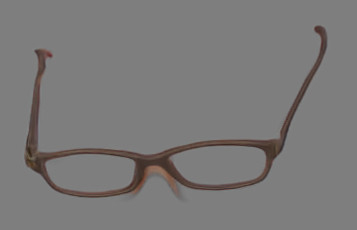}\\
    	all
    \end{subfigure}
	\begin{subfigure}[t]{0.22\columnwidth}
    	\centering
    	\includegraphics[width=\columnwidth]{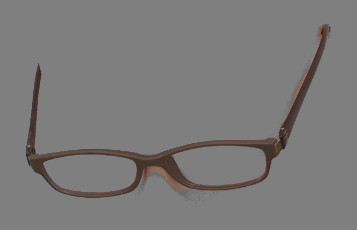}\\
    	ground truth
    \end{subfigure}
	\caption{
	Differences depending on where the model is being fit. The shape is best fit under $\mathbf{w}$, although the texture does not match, and better overall reconstruction is achieved when all network parameters are fine-tuned.
	}
   \label{fig:variables_to_fit}
\end{figure}

\begin{table*}[t]
\centering
\scriptsize		
\begin{tabular}{@{\extracolsep{2pt}}l|cc|cccc|cccc} 
\toprule

\multicolumn{1}{c}{} & \multicolumn{2}{c}{ DNR~\cite{Thies2019NeuralTextures} } & \multicolumn{4}{c}{ Ours }  & \multicolumn{4}{c}{ NeRF~\cite{mildenhall2020nerf} } \\

\multicolumn{1}{c}{} & \multicolumn{2}{c}{ trained from scratch } & \multicolumn{4}{c}{ finetuning category model }  & \multicolumn{4}{c}{ trained from scratch } \\
\cmidrule(lr){2-3} \cmidrule(lr){4-7} \cmidrule(lr){8-11}
 & N=30  & N=100 & N=3 & N=10 & N=30 & N=100 & N=3 & N=10 & N=30 & N=100 \\
\hline
$\text{PSNR}$ &  38.75 & 40.05 & 36.53 & 39.35 & 41.61 & 43.42 & 31.20 & 37.21 & 43.32 & $\mathbf{45.28}$ \\
$\text{PSNR}_M$ & 21.48 & 22.43 & 19.01 & 21.78 & 24.00& 25.80 & 15.41 & 21.25 & 27.49 & $\mathbf{29.80}$ \\
SSIM & 0.9858 & 0.9897 & 0.9824 & 0.9890 & 0.9921 & 0.9942 & 0.9600 & 0.9845 & 0.9947 & 0.9962  \\
Mask IoU & 0.9293 & 0.9407 & 0.8864 & 0.9350 & 0.9585 & 0.9682 & N/A & N/A & N/A & N/A \\

\bottomrule
\end{tabular}\caption{
Reconstruction results with varying numbers of input images $N$ for unseen instances, for the DNR baseline without the category model, finetuning our category-level model, and NeRF. Fine-tuning the category model provides similar quality to DNR with $>3\times$ fewer input views, and provides $\sim 3$ dB improvement with the same number of input views. NeRF generates better results with $N\geq30$ views, but is significantly slower to train and render novel views. }
\label{tab:few_shot_num_images}
\end{table*}

In Table~\ref{tab:ablation_study}, we quantitatively evaluate reconstructions of $5$ unseen instances using only $N=3$ input images, by fine-tuning all network parameters together with the latent code, and show qualitative results in Figure~\ref{fig:few_shot_reconstruction_comparison}. 
We use the same baselines as in Section~\ref{sec:view-interpolation}, and report statistics across the $5$ instances. We halt the optimization at 1000 steps, because running the optimization to convergence overfits to only the visible data, reducing the performance on unseen views. We observe that the VAE model is inferior, and that stacking the proxy inputs in our model performs better compared to z-buffering in DNR, because the eyeglasses' arms can be occluded by the front proxy, preventing the optimization of the side textured proxy. Figure~\ref{fig:few_shot_reconstruction} shows the input images and reconstructed views using our model, illustrating accurate reproduction of view-dependent effects on the bridge and novel views from an unseen side of the glasses.

\begin{figure}[t]
	\centering
    \setlength{\tabcolsep}{0.0em} %
    \begin{tabular}{l m{0.94\columnwidth}}
    Ours~ &\includegraphics[width=0.94\columnwidth]{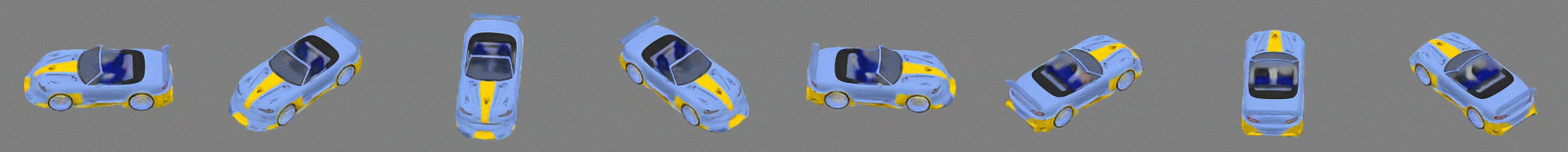}\\
    \vspace{-14pt}\\
    GT~ &\includegraphics[width=0.94\columnwidth]{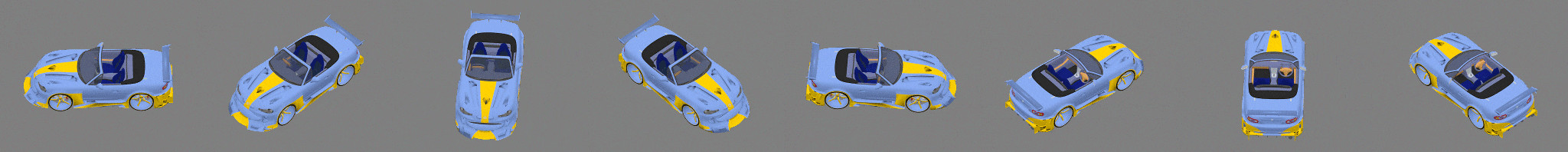}\\
    Ours~ &\includegraphics[width=0.94\columnwidth]{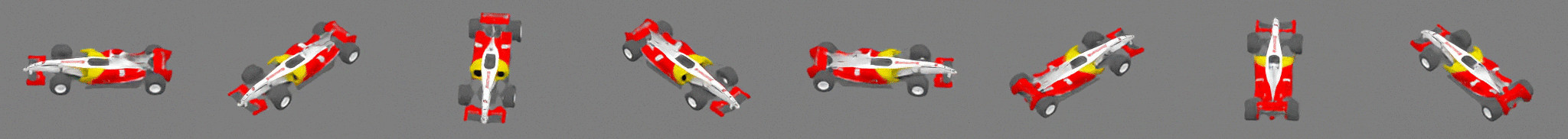}\\
    \vspace{-14pt}\\
    GT~ &\includegraphics[width=0.94\columnwidth]{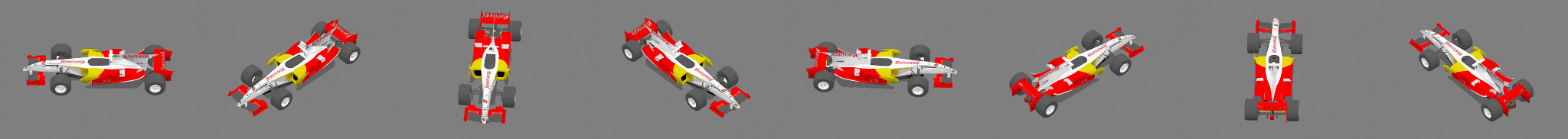}\\
    Ours~ &\includegraphics[width=0.94\columnwidth]{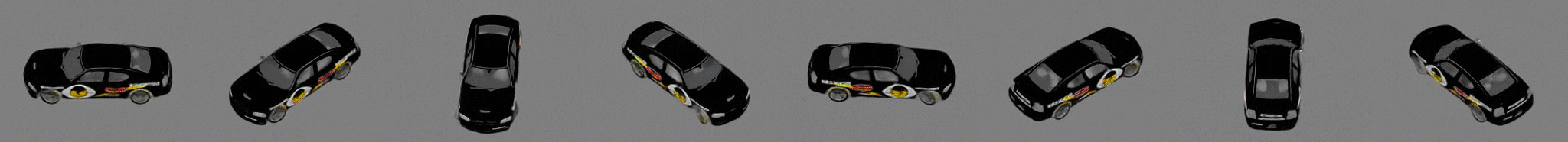}\\
    \vspace{-14pt}\\
    GT~ &\includegraphics[width=0.94\columnwidth]{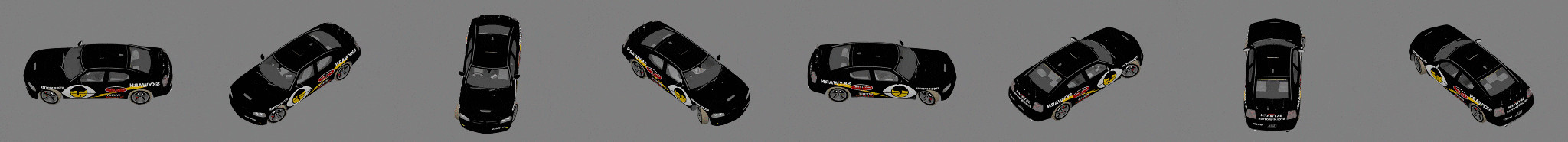}\\
    \end{tabular}

	\caption{
	Reconstruction results on ShapeNet cars using textured proxies based on AtlasNet reconstructions. See supplementary video for more results.
	}
   \label{fig:cars_reconstruction}
\end{figure}

To demonstrate the power of our representation, we compare reconstructions of unseen objects with increasing number of input images $N$, using our GeLaTO, and the DNR baseline described in Section~\ref{sec:view-interpolation}, that is exclusively trained on the unseen instance. Similar to Thies~\etal~\cite{Thies2019NeuralTextures}, we optimize the neural texture for 30k and 100k steps for $N=30$ and $N=100$ respectively. 
We also compare with Neural Radiance Fields (NeRF)~\cite{mildenhall2020nerf}, a concurrent novel-view synthesis technique that uses a volumetric approach that does not require proxy geometry. Table~\ref{tab:few_shot_num_images} and Figure~\ref{fig:few_shot_reconstruction} show that our representation achieves better results than the DNR baseline with more than $3\times$ less input images. Using the same number of input images, our reconstructions have PSNR score $\sim 3$ dB higher than the model trained from scratch. Compared to NeRF, our model is more accurate with few views, although NeRF is significantly better with denser sampling. Moreover, training the DNR baseline takes 50 and 150 minutes on 15 GPUs for $N=30$ and $N=100$ respectively, whereas fine-tuning GeLaTO takes less than 4 minutes on a single GPU. Training NeRF takes $4$ hours on 4 GPUs and rendering a single using NeRF takes several seconds, making it unsuitable for real-time rendering, while DNR and GeLaTO render new views under 20ms on a NVidia 1080 Ti.

\begin{table*}[t]
\small		
 
\centering
\setlength{\tabcolsep}{3pt}

\begin{tabular}{l|cccc}
\toprule
Fit variables & $z$ & $w$ & texture & all \\
\hline
$\text{PSNR}$ & 31.30  & 36.50  & 37.12  & \textbf{37.19} \\
$\text{PSNR}_M$ & 13.85 & 18.85 & 19.59 & \textbf{19.64} \\
SSIM & 0.9638 & 0.9833 & 0.9841 & \textbf{0.9842} \\
Mask IoU & 0.7242 & \textbf{0.9152} & 0.8984 & 0.9012 \\
\bottomrule
\end{tabular}
\caption{
Comparison of reconstructions when fitting in different spaces.
$z$ is the instance latent code, $w$ is the transformed latent code, texture refers to fitting also the parameters of the texture generators, and all refers to fine-tuning the neural rendering network as well.
}
\label{tab:variables_to_fit}
\end{table*}

Finally, we evaluate the choice of which variables to optimize during few-shot reconstruction in Table~\ref{tab:variables_to_fit}, and show comparative qualitative results in Figure~\ref{fig:variables_to_fit}. Optimizing the transformed latent code $\mathbf{w}$ reconstructs the shape best as measured by the mask IoU, albeit with a strong color mismatch. Fine-tuning all the network parameters generates the best results as measured by PSNR.

\subsection{Results on ShapeNet}

We show results of modeling ShapeNet cars using textured proxies based on AtlasNet reconstructions. We train a model on $100$ car instances using $500$ views. We use $5$ textured proxies, with a $128 \times 128$ resolution each, and increase the first layer of the neural renderer from $32$ to $64$ channels to accommodate the extra proxies' channels. Figure~\ref{fig:cars_reconstruction} shows unseen view reconstruction results, scoring a PSNR of $30.99$ dB on a held-out set.

\begin{figure}[t]
	\centering
	\includegraphics[height=1.25cm]{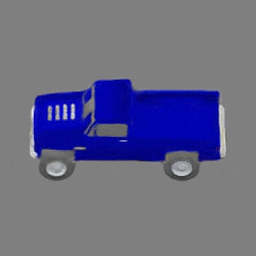}\hspace{2pt}\includegraphics[height=1.25cm]{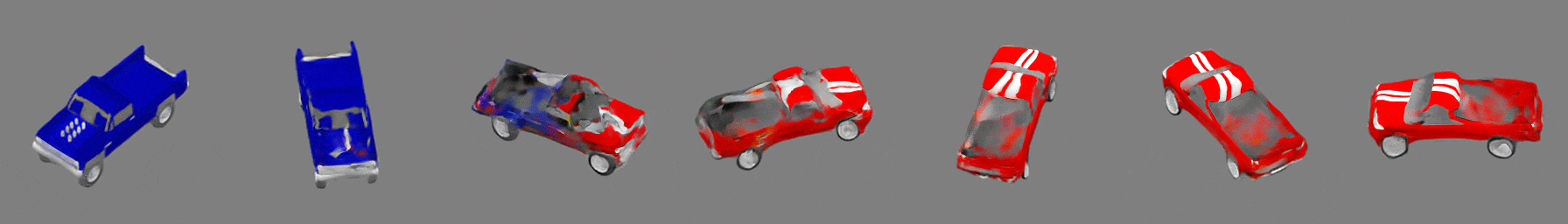}\hspace{2pt}\includegraphics[height=1.25cm]{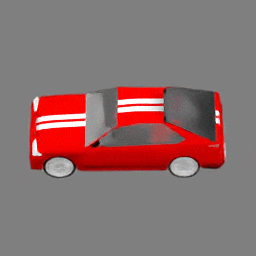}\\
	\includegraphics[height=1.25cm]{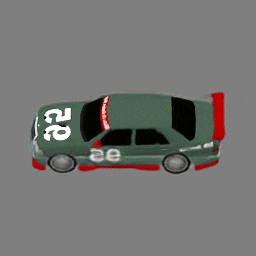}\hspace{2pt}\includegraphics[height=1.25cm]{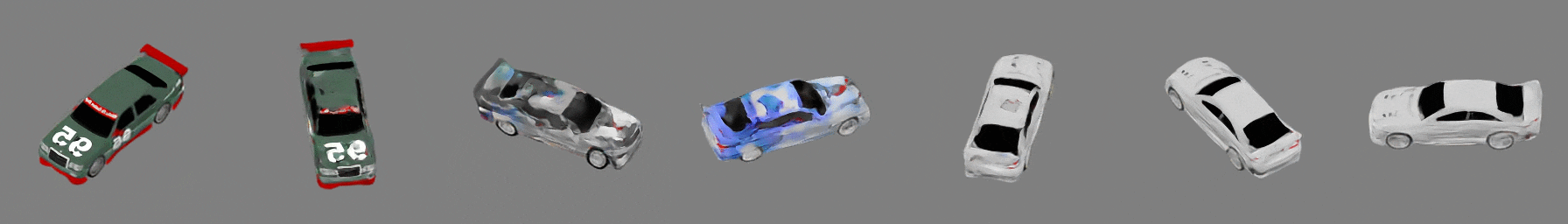}\hspace{2pt}\includegraphics[height=1.25cm]{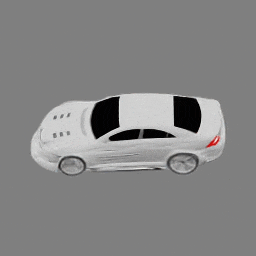}\\
	\includegraphics[height=1.25cm]{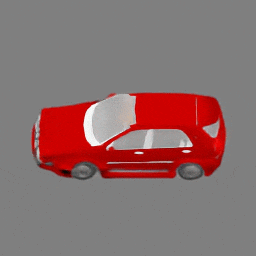}\hspace{2pt}\includegraphics[height=1.25cm]{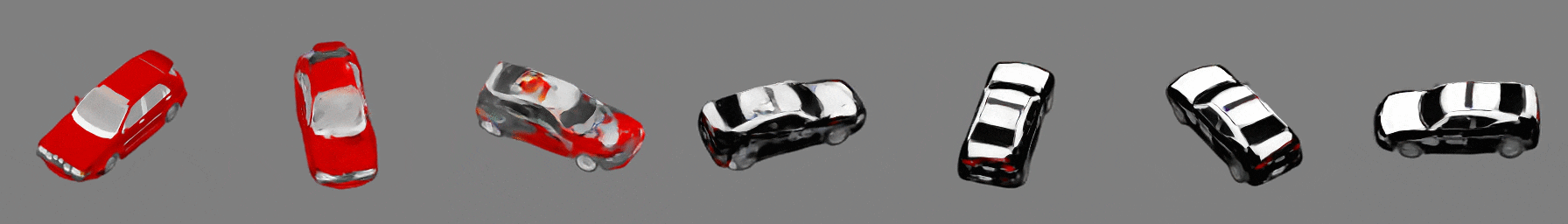}\hspace{2pt}\includegraphics[height=1.25cm]{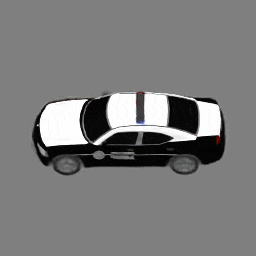}\\
	\includegraphics[height=1.25cm]{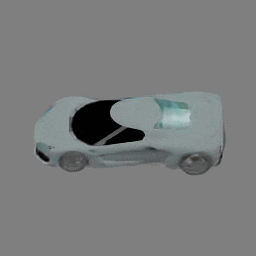}\hspace{2pt}\includegraphics[height=1.25cm]{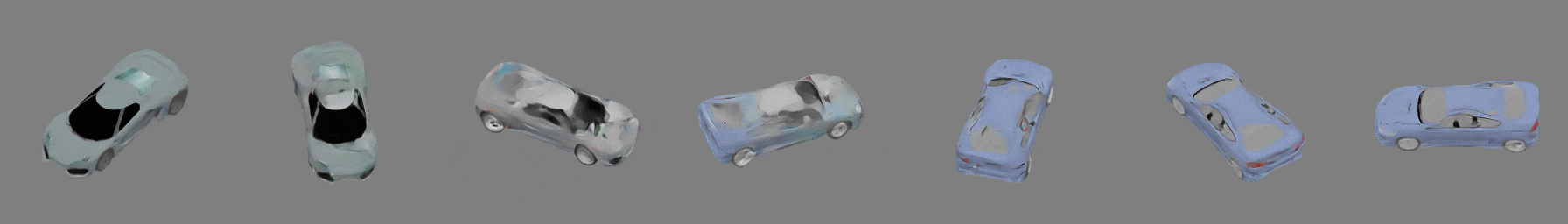}\hspace{2pt}\includegraphics[height=1.25cm]{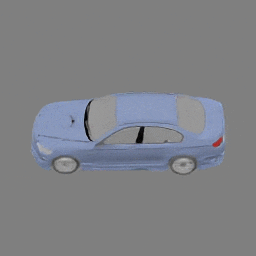}\\

	\caption{
	Instance interpolations on ShapeNet. Left: reconstructed view of start instance. Middle: latent texture code interpolation while keeping proxy geometry constant. Right: target instance reconstruction using its proxy geometry.
	}
   \label{fig:car_interpolations}
\end{figure}

Figure~\ref{fig:car_interpolations} shows smooth latent interpolation of the latent code of the textured proxies while maintaining the proxy geometry of the first car. Although the proxy geometry is different between instances, Groueix~\etal~\cite{Groueix2018AtlasNet} observe that the semantically similar areas of the car are modeled consistently by the same parts of the AtlasNet patches, allowing our model to generate plausible renderings when modifying only the neural texture. Using the proxy geometry of the first instance creates some artifacts, like the white stripes on the first example that are tilted compared to the car's main axis.
The eyeglasses interpolation results are more realistic due to a smaller degree of variability in the object class. Please see the supplementary video for more results.

\subsection{Neural textures}

\begin{figure}[t]
    \hspace{-30pt}
	\centering
	\begin{subfigure}[t]{0.48\columnwidth}
    	\centering
    	\includegraphics[height=2.8cm]{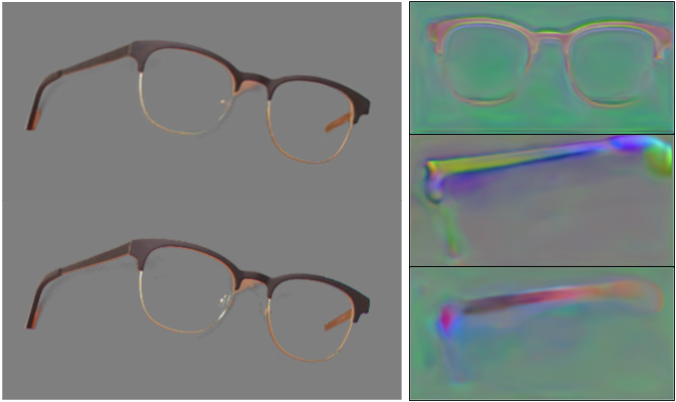}\\
    \end{subfigure}
	\begin{subfigure}[t]{0.48\columnwidth}
    	\centering
    	\includegraphics[height=2.8cm]{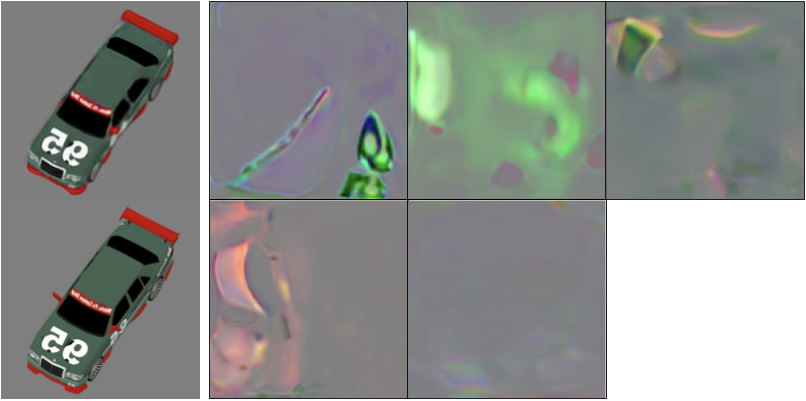}%
    \end{subfigure}
	\caption{
	Learnt neural textures for eyeglasses and cars. Left top: reconstructed view, left bottom: ground truth, right: neural textures. Note the high frequency details encoding the eyeglasses' shape and the number decal on the car.
	}
   \label{fig:neural_textures}
\end{figure}

We visualize the learned neural textures in Figure~\ref{fig:neural_textures}, showing the first three channels as red, green and blue. They contain high frequency details of the object, such as the eyeglasses shape and decals on the car.

\subsection{Limitations}

Our model has several limitations. When seen from the side, planar proxies almost disappear when rasterized to the target view, creating artifacts on the eyeglasses arms in view interpolations, as seen for a few instances in the supplementary video. Another type of artifacts stems from inaccurate matting in the captured dataset, as seen by the remaining skin color shadows in row 4 of Figure~\ref{fig:view_interpolation_comparison} and the incomplete transparent eyeframe in row 6. In the case of few-shot reconstruction, a major limitation of our model is the requirement of known pose and proxy geometry, which can be tackled as a general 6D pose estimation in the case of planar billboard proxies.
\section{Application: Virtual Try-On}

\begin{figure}[t]
    \centering
    \setlength{\tabcolsep}{4pt}
    \begin{tabular}{ccc}
	\includegraphics[height=3.7cm]{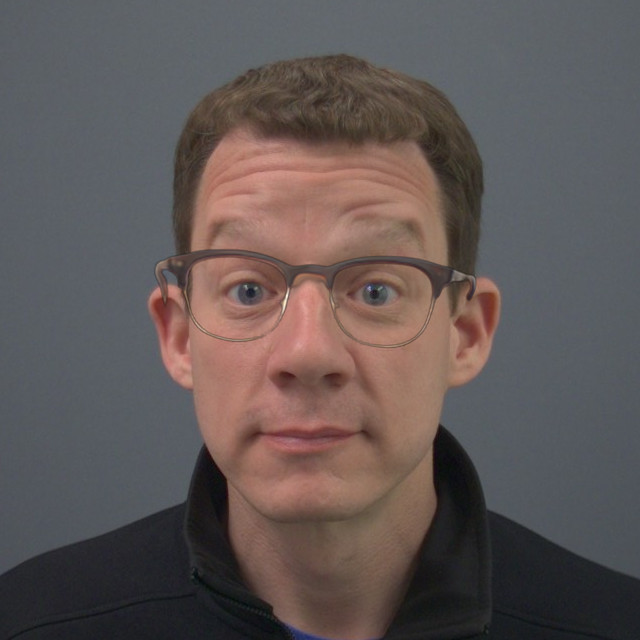} &
	\includegraphics[height=3.7cm]{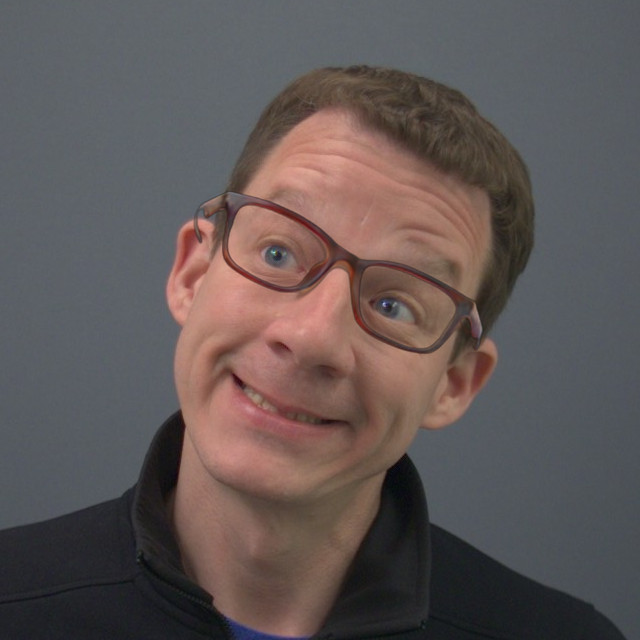} &
	\includegraphics[height=3.7cm]{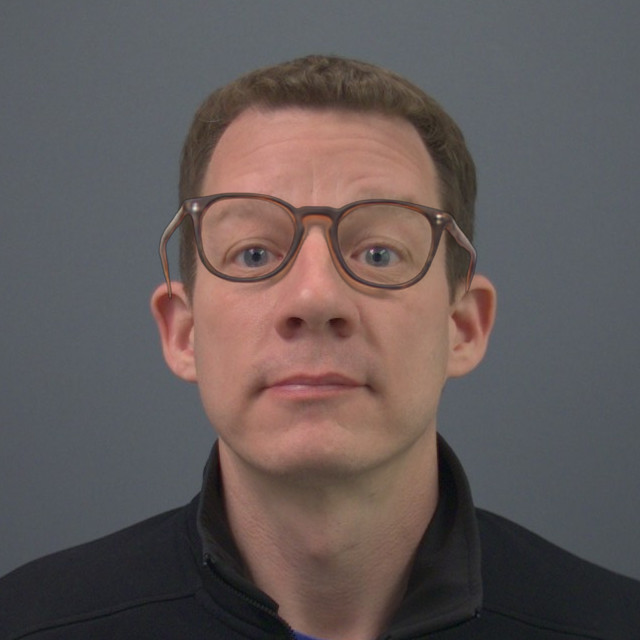}\\
    \end{tabular}
	\caption{
	Virtual try-on application for eyeglasses frames, in which a user without eyewear can virtually place reconstructed glasses on themselves. The eyeglasses are generated by our model given the user's head pose, and composited on the user's view. See supplementary video for more results.}
   \label{fig:virtual_try_on}
\end{figure}

Our generative model of eyeglasses frames can enable the experience of virtually trying-on a pair of eyeglasses~\cite{Zhang2017GlassesTryOn}. Additionally, the learned latent space allows a user to modify the appearance and shape of eyeglasses by modifying the input latent code. We prototype such a system in Figure~\ref{fig:virtual_try_on}, where we capture a video of a user at close distance who is not wearing eyewear, track their head pose using~\cite{GoogleFacemesh}, place the textured proxies on the head frame of reference, render the neural proxies to into a RGBA eyeglasses layer and finally composite it onto the frame. Our neural renderer network is sufficiently lightweight -- running under $20$ms on a NVidia 1080Ti -- that such a system could be made to run interactively.

\section{Conclusion}

We present a novel compact and efficient representation for jointly modeling shape and appearance. Our approach uses coarse proxy geometry and generative latent textures. We show that by jointly modeling an object collection, we can perform latent interpolations between seen instances, and reconstruct unseen instances at high quality with as few as 3 input images. We show results on a dataset consisting of real images and alpha mattes of eyeglasses frames, containing strong view-dependent effects and semi-transparent materials, and on ShapeNet cars. The current approach assumes known proxy geometry and pose; modeling the distribution of proxy geometry and estimating both its parameters and pose on a given image remains as future work.

\clearpage

\bibliographystyle{splncs04}
\bibliography{egbib}
\end{document}